\documentclass{l4dc2026}
\makeatletter
\renewcommand\ps@jmlrtps{%
  \let\@mkboth\@gobbletwo
  \def\@oddhead{}%
  \let\@evenhead\@oddhead
  \def\@oddfoot{\@titlefoot}%
  \let\@evenfoot\@oddfoot
}
\makeatother

\title[Efficient Natural Language to Temporal Logic Translation]{Efficient Natural Language to Temporal Logic Translation\\for Real-Time Physical AI}
\usepackage{times}
\usepackage{booktabs}
\usepackage{longtable}
\usepackage{array}
\usepackage{xcolor}
\usepackage{tikz}
\usetikzlibrary{arrows.meta, decorations.pathreplacing, shadows, backgrounds}
\definecolor{mypurple}{HTML}{C701FF}
\definecolor{mymidblue}{HTML}{008DFF}
\definecolor{myorange}{HTML}{FF9D3A}
\definecolor{mygreen}{HTML}{4ECB8D}
\definecolor{myred}{HTML}{D83034}

\newcommand{\labeling}{L}
\newcommand{\R}{\mathbb{R}}
\newcommand{\Prob}{\mathbb{P}}
\newcommand{\U}{\mathbb{U}}
\newcommand{\X}{\mathbb{X}}

\DeclareMathOperator{\AP}{AP}
\newcommand{\Frac}{\mathcal{F}}
\newcommand{\LTL}{\mathrm{LTL}}
\newcommand{\instr}{\texttt{i}}
\newcommand{\Instr}{\texttt{I}}
\newcommand{\opt}{o}

\newcommand{\Rpi}{R\ensuremath{\pi}}
\newcommand{\toolarge}{{too large}}

\newcommand{\equivL}{\equiv_{\mathcal{L}}}
\newcommand{\word}{\boldsymbol{\omega}}

\newcommand{\tool}[1]{\textsc{#1}}
\usepackage{xspace}
\usepackage{wrapfig}
\newcommand{\laya}{\tool{Laya}\xspace}
\newcommand{\jev}{\tool{Jev}\xspace}

\newif\ifsegshadow
\makeatletter
\newcommand{\seg@outer}{1.5pt}%
\newcommand{\seg@inner}{0.25em}%
\newcommand{\seg@piece}[5]{%
  \tikz[baseline=(n.base)]{%
    \node[inner xsep=0pt, inner ysep=1pt, outer sep=0pt, text height=0.68em, text depth=0.22em, text=#1,
      font=\ttfamily\frenchspacing] (n)
      {\hspace*{\ifnum#3=1 \seg@outer\else\seg@inner\fi}#5\hspace*{\ifnum#4=1 \seg@outer\else\seg@inner\fi}};
    \def\seg@r{1.5pt}%
    \def\seg@left{\ifnum#3=1 -- ([xshift=\seg@r]n.north west) arc[start angle=90, end angle=180, radius=\seg@r]
      -- ([yshift=\seg@r]n.south west) arc[start angle=180, end angle=270, radius=\seg@r]
      \else -- (n.north west) -- (n.south west)\fi}%
    \def\seg@right{\ifnum#4=1 -- ([xshift=-\seg@r]n.south east) arc[start angle=270, end angle=360, radius=\seg@r]
      -- ([yshift=-\seg@r]n.north east) arc[start angle=0, end angle=90, radius=\seg@r]
      \else -- ([xshift=0.4pt]n.south east) -- ([xshift=0.4pt]n.north east)\fi}%
    \begin{pgfonlayer}{background}
      \ifsegshadow\fill[overlay, black!25, transform canvas={shift={(0.5ex,-0.5ex)}}] (n.north) \seg@left -- (n.south)
        \seg@right -- cycle;\fi
      \fill[overlay, #2] (n.north) \seg@left -- (n.south) \seg@right -- cycle;
    \end{pgfonlayer}}}
\makeatother
\ExplSyntaxOn
\seq_new:N \l__seg_words_seq
\int_new:N \l__seg_n_int
\int_new:N \l__seg_i_int
\NewDocumentCommand{\segbox}{O{black} m m m}
  {
    \seq_set_split:Nnn \l__seg_words_seq { ~ } { #4 }
    \seq_remove_all:Nn \l__seg_words_seq { }
    \int_set:Nn \l__seg_n_int { \seq_count:N \l__seg_words_seq }
    \int_zero:N \l__seg_i_int
    \seq_map_inline:Nn \l__seg_words_seq
      {
        \int_incr:N \l__seg_i_int
        \int_compare:nNnF { \l__seg_i_int } = { 1 } { \allowbreak }
        \use:c { seg@piece } {#1} {#3}
          { \int_compare:nNnTF { \l__seg_i_int } = { 1 } { 1 } { 0 } }
          { \int_compare:nNnTF { \l__seg_i_int } = { \l__seg_n_int } { 1 } { 0 } }
          { ##1 }
      }
  }
\ExplSyntaxOff
\newcommand{\segtype}[1]{\segbox[myorange!50!black]{}{myorange!20}{#1}}
\newcommand{\segquestion}[1]{\segbox[mygreen!50!black]{}{mygreen!15}{#1}}
\newcommand{\segopt}[1]{\segbox[mypurple!50!black]{}{mypurple!12}{#1}}
\newcommand{\seginstr}[1]{\segbox[mymidblue!50!black]{}{mymidblue!12}{#1}}
\newcommand{\segsp}[1]{\segbox[black!50!white]{}{gray!15}{\kern-0.15em#1\kern-0.15em}}

\author{%
 \Name{Oliver Schön} \Email{oschoen@ethz.ch}\\
 \Name{Adrien Banse} \Email{abanse@ethz.ch}\\
 \Name{John Lygeros} \Email{jlygeros@ethz.ch}\\
 \Name{Lars Lindemann} \Email{llindemann@ethz.ch}\\
 \addr Automatic Control Laboratory, ETH Zürich, Switzerland%
}

\begin{document}

\maketitle

\begin{abstract}%
    Robots can be controlled to perform tasks with guaranteed correctness when the task instructions are stated in a formal language. 
    Large language models (LLMs) have been widely used to allow users to provide such instructions in natural language (NL), by translating NL into formal language. When deploying such translators on a robot, however, we face limited computing and energy resources, while offloading computation or data to the cloud may be infeasible or undesirable. Yet an LLM is a general-purpose text generator, whereas a formula in a fixed formal language fragment follows a well-defined syntax.
    We therefore ask whether text generation is needed at all.
    For the formal language of linear temporal logic (LTL), we show that this translation can instead be broken down into answering a set of questions with pre-defined answers.
    We then construct a System-One model, a recent class of NL-based classifier, to provide such answers and enable us to translate a natural language instruction into an LTL instruction.
    On three benchmarks, our translator matches or exceeds the accuracy of the best local LLMs available, at up to one order of magnitude lower runtime and energy consumption. We show that it generalizes to unseen atomic propositions and temporal patterns.
\end{abstract}


\section{Introduction}
\label{sec:intro}

Symbolic control methods compute, from a model of a system and a task description, a controller that carries out the task
by construction \citep{belta2017}. This guarantee has a price: the task must be stated in a
formal language, such as a temporal logic. Consider a mobile robot that receives
the instruction \emph{``Visit street 5 while ensuring you never enter parking lot 2''}. In \emph{linear temporal
logic} (LTL), it reads
``$\Diamond\, \texttt{street\_5} \;\wedge\; \Box\, \neg\, \texttt{parking\_lot\_2}$'',
where the atomic propositions \texttt{street\_5} and \texttt{parking\_lot\_2} hold when the robot is at the
corresponding location. Writing such formulas takes expertise that most users do not have and is, together
with scalability, one of the two main bottlenecks in applying formal methods \citep{rozier2016}. This makes
automatic translation of natural language (NL) into a formal language important.

Many recent translators from natural to formal language rely on the most powerful language models available,
from \tool{Codex} to \tool{GPT-5}, \tool{Claude Opus}, and the latest \tool{Gemini} models
\citep{cosler2023,chen2024,wang2025,ma2025,bayat2026,newcomb2026}. Such translators have outperformed earlier translation methods on
shared benchmarks \citep{pan2023,chen2023,liu2023} and are therefore considered the state of the art
(Appendix~\ref{app:related_work} provides a literature review).
Most of these models are proprietary and behind an API, or too large
to be run on a single computer, and only a minority of translators run entirely on local, open-weight models.
On a robot, however, the translator must run onboard. New instructions may arrive during operation and require translating and replanning in real time,
with limited compute and energy resources, and often without access to the cloud, e.g., for security or reliability reasons
\citep{kehoe2015,firoozi2025}. 

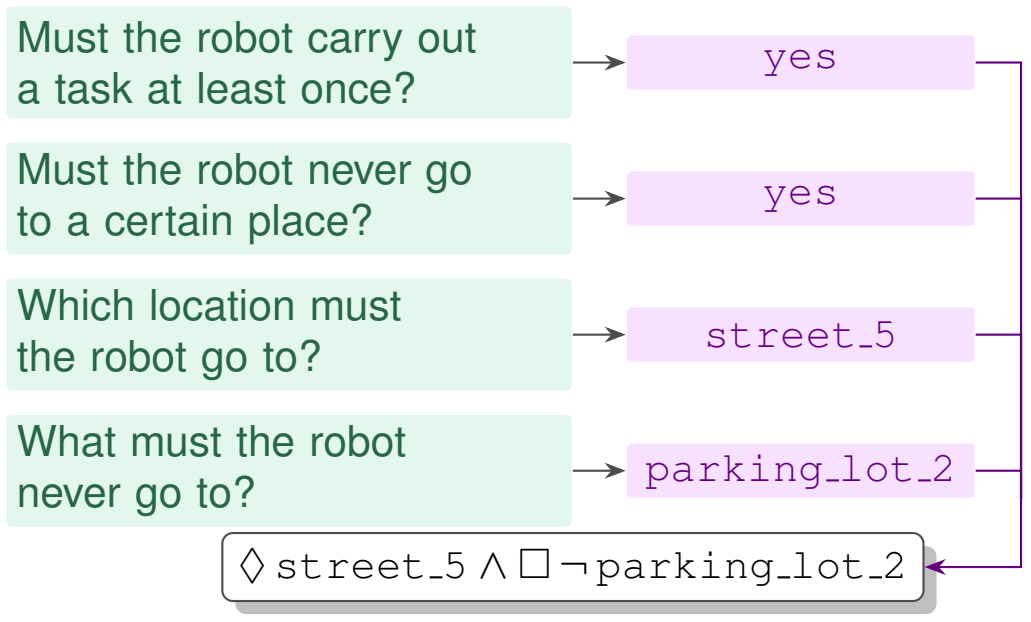
\begin{wrapfigure}{r}{72mm}
\vspace{-0mm}
\centering
\begin{tikzpicture}[x=1mm, y=1mm, font=\scriptsize\sffamily,
    que/.style={rounded corners=1pt, fill=mygreen!15, text=mygreen!50!black, inner xsep=2pt, inner ysep=1.5pt,
      text width=36mm, minimum height=7.4mm, align=left, anchor=west},
    ans/.style={rounded corners=1pt, fill=mypurple!12, text=mypurple!50!black, inner sep=1pt, minimum width=23mm,
      minimum height=3.6mm, align=center, anchor=west, font=\scriptsize\ttfamily},
    arr/.style={draw, -{Stealth[length=4pt]}, gray!60!black, thin},
  ]
  \foreach \q/\a [count=\k] in {
      {Must the robot carry out\\a task at least once?}/{yes},
      {Must the robot never go\\to a certain place?}/{yes},
      {Which location must\\the robot go to?}/{street\_5},
      {What must the robot\\never go to?}/{parking\_lot\_2}} {
    \node[que] (q\k) at (0, -9*\k) {\q};
    \node[ans] (a\k) at ([xshift=3.5mm] q\k.east) {\a};
    \draw[arr] (q\k.east) -- (a\k.west);
  }
  \coordinate (join) at ([xshift=3mm] a1.east |- a4.south);
  \path (join) ++(0, -4.5) coordinate (fy);
  \node[draw=gray!60!black, thin, fill=white, rounded corners=2pt, drop shadow, inner sep=3pt] (f)
    at (q4.east |- fy) {$\Diamond\,\texttt{street\_5} \wedge \Box\,\neg\,\texttt{parking\_lot\_2}$};
  \foreach \k in {1, ..., 4} \draw[thin, mypurple!50!black] (a\k.east) -| (join);
  \draw[arr, mypurple!50!black] (join) |- (f.east);
\end{tikzpicture}
\vspace{-1mm}
\caption{Questions of the running example.}\label{fig:questions}
\vspace{-0mm}
\end{wrapfigure}
We ask whether such generative models are needed at all for physical AI applications. Indeed,  translators can identify a formula by asking a model a few simple questions; for the running example, four questions suffice (Figure~\ref{fig:questions}). The two \emph{``yes''} answers select the operators $\Diamond$ (eventually) and $\Box\neg$ (never), and the two choices fill them with
\texttt{street\_5} and \texttt{parking\_lot\_2}. Recently, general-purpose frontier classifiers were introduced with TypeSafe's model \jev
\citep{jev2026}. These recent models read unstructured text together with a question and return a probability for each of its possible answers, instead of generating text. These \emph{System-One models} are particularly cheap and fast: they answer a
question in a single forward pass instead of one per generated token, and the open-weight model \laya \citep{laya2026} has
421 million parameters, orders of magnitude fewer than state-of-the-art LLMs and small enough to run locally on a robot.

\paragraph{Contributions.}
We introduce the first family of translators from NL into LTL built on a System-One model rather than on a generative
language model. Its questions are designed for each system's fragment of possible requirements and atomic propositions, so the same
architecture serves different control systems (Section~\ref{sec:method}). By construction, its number of forward passes per instruction is bounded in
advance, and every output is a well-formed formula; generative
models lack both properties.
We compare our translators with 20 off-the-shelf and 5 fine-tuned local language models on three benchmarks, including a new
Mine Rover benchmark, measuring accuracy, runtime, and energy consumption on a laptop and a Raspberry Pi~5
(Section~\ref{sec:experiments}); to our knowledge, no prior study of translation into temporal logic measures energy.
Their accuracy matches or exceeds that of language models with the same characteristics, i.e., local
and fine-tuned on the same data, at a much lower cost.
Beyond that, our translators generalize to atomic propositions and temporal patterns unseen in training and, on the Mine
Rover, make 3.6 times fewer errors than the most accurate fine-tuned language model at an order of magnitude lower
runtime and energy (Section~\ref{sec:results}).
This is the first approach that enables efficient NL to formal language translation for real-time physical AI that uses embedded hardware such as a Raspberry Pi 5.

\section{Problem Formulation}
\label{sec:prelim}
We consider discrete-time control systems $
  x_{t+1} = f(x_t, u_t)$, with $x_t \in \X$, $u_t \in \U$, $t \in \mathbb{N}$,
where $x_{t+1} \in \X$ is the next state and $f\colon \X \times \U \to \X$ the transition function. A \emph{controller}
$C\colon \X^+ \to \U$ maps finite nonempty sequences of states to inputs: from an initial state $x_0 \in \X$, it
generates the closed-loop trajectory $\mathbf{x} = x_0 x_1 x_2 \cdots$ with $x_{t+1} = f(x_t, C(x_0 x_1 \cdots x_t))$
for all $t \in \mathbb{N}$.

\subsection{Linear Temporal Logic}
\label{sec:prelim-ltl}
Let $\AP$ be a finite set of atomic propositions. Formulas of \emph{linear temporal logic} (LTL) over $\AP$ \citep{baier2008} are generated by
the grammar
\[
  \varphi \;::=\; \top \;\mid\; p \;\mid\; \neg\varphi \;\mid\; \varphi \wedge \varphi \;\mid\; \bigcirc\varphi
  \;\mid\; \varphi \,\mathcal{U}\, \varphi, \qquad p \in \AP,
\]
with the usual derived operators $\vee$, $\rightarrow$, $\leftrightarrow$, $\Diamond$ (eventually), $\Box$
(always), $\mathcal{R}$ (release), and $\mathcal{W}$ (weak until). A formula is evaluated on an infinite sequence $\word = \sigma_0\sigma_1\sigma_2\cdots$, where $\sigma_t \subseteq
\AP$ is the set of atomic propositions that hold at time $t$; we write $\word \models \varphi$ if $\word$ satisfies
$\varphi$ \citep[see][for the formal semantics]{baier2008}. Informally, $\bigcirc\varphi$ requires $\varphi$ at the next
time, $\varphi \,\mathcal{U}\, \psi$ requires $\psi$ at some time and $\varphi$ until then, $\Diamond\varphi$ requires
$\varphi$ at some time, and $\Box\varphi$ at all times. The \emph{language} $\mathcal{L}(\varphi) = \{\word \colon \word
\models \varphi\}$ is the set of sequences that satisfy $\varphi$. We write $\LTL(\AP)$ for the set of all LTL formulas over $\AP$. A \emph{fragment} of LTL over $\AP$ is a
subset $\Frac(\AP) \subseteq \LTL(\AP)$. The choice of atomic propositions depends on what the system can observe and do,
and the corresponding fragment contains the formulas of the tasks it may carry out. Two formulas $\varphi$ and $\psi$ are \emph{(language) equivalent}, written $\varphi \equivL \psi$, if $\mathcal{L}(\varphi) = \mathcal{L}(\psi)$.
Equivalence is our criterion of correctness: a translation is correct if it is equivalent to the intended formula (in a data set, the \emph{reference} formula).

The atomic propositions link LTL to the executions of the control system: a \emph{labeling function} $\labeling\colon \X \to
2^{\AP}$ assigns to every state of the system's state space $\X$ the atomic propositions that hold in it, so that a
trajectory $\mathbf{x}=x_0 x_1 x_2 \cdots$ of the system produces the sequence $\word =\labeling(x_0)\labeling(x_1)\labeling(x_2)\cdots$.
Given a formula $\varphi$, a controller can be computed under which every trajectory of the closed-loop system from a given set
of initial states satisfies $\varphi$ \citep{belta2017}.

\subsection{Problem Statement}
\label{sec:problem}

A user states a task for the control system as an instruction $\instr\in\mathbb{I}$, where $\mathbb{I}$ is
the set of finite strings of natural language.
A \emph{translator} $T$ maps every instruction $\instr$ to an output $T(\instr)$, which is a finite string of symbols. The output is
\emph{syntactically correct} if it is a valid LTL formula (in the syntax of Section~\ref{sec:prelim-ltl}), i.e.,
$T(\instr) \in \LTL(\AP)$. Note that a syntactically correct formula may still differ in meaning from the intended formula, and it may lie outside the fragment
$\Frac(\AP)$ of formulas sensible for the system at hand. We model the instructions that the system receives, together with the formulas that their users
intend, as a random variable $(\Instr, \Phi)$ with values in $\mathbb{I} \times \Frac(\AP)$. The \emph{accuracy} of the translator $T$ is
the probability $\Prob(T(\Instr) \equivL \Phi)$ that its output is equivalent to the intended formula. An output that
is not syntactically correct is never equivalent. The \emph{cost} of a translator $T$ is the runtime and
energy it consumes per instruction on a given hardware.

Given a control system with atomic propositions $\AP$ and an LTL fragment $\Frac(\AP)$, our problem is to find a
translator $T$ that works without a network connection and without asking the user for clarification, with an accuracy
comparable to, and a cost lower than, those of existing LLM baselines. Such a translator is particularly suitable for
physical AI applications, where instructions must be translated on the spot by the hardware onboard the system.

\section{System-One Translators}
\label{sec:method}
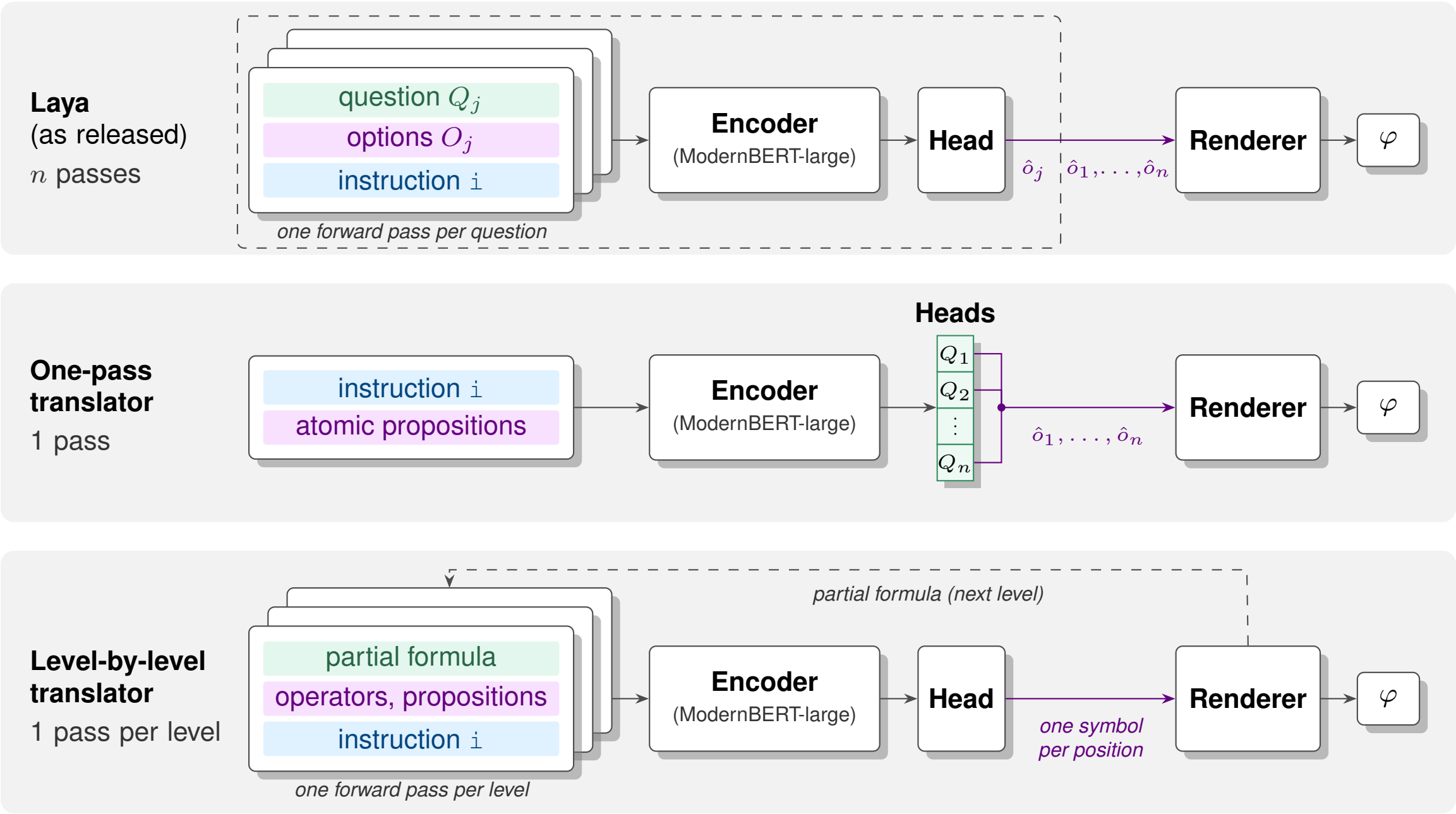
\begin{figure}[htb]
\centering
\begin{tikzpicture}[x=1mm, y=1mm,
    font=\scriptsize\sffamily,
    box/.style={draw=gray!60!black, thin, fill=white, rounded corners=2pt, drop shadow, align=center, inner sep=3pt},
    card/.style={draw=gray!60!black, thin, fill=white, rounded corners=2pt, drop shadow},
    chip/.style={rounded corners=1pt, minimum width=31mm, minimum height=3.6mm, inner sep=1pt},
    chq/.style={chip, fill=mygreen!15, text=mygreen!50!black},
    cho/.style={chip, fill=mypurple!12, text=mypurple!50!black},
    chi/.style={chip, fill=mymidblue!12, text=mymidblue!50!black},
    arr/.style={draw, -{Stealth[length=4pt]}, gray!60!black, thin},
    lab/.style={text=gray!40!black, align=center, font=\tiny\sffamily\itshape},
    sel/.style={lab, text=mypurple!50!black},
    title/.style={anchor=west, align=left, text width=21.5mm, inner sep=0pt},
    enc/.style={box, minimum height=11mm, text width=22mm},
    head/.style={box, minimum height=11mm, text width=7mm, anchor=west},
    ren/.style={box, minimum height=11mm, text width=13mm, anchor=west},
    outp/.style={box, minimum height=5.5mm, minimum width=6.5mm, inner sep=1pt, anchor=west},
  ]
  \def\pl{0}\def\pr{152.4}%
  \def\cx{43}%
  \def\cw{17}%
  \def\ex{80}%
  \def\hx{98}%
  \def\hy{96}%
  \def\rx{123}%
  \def\ox{142}%
  \def\dl{2}%
  \def\cards#1{%
    \foreach \k in {2, 1} \draw[card] (\cx-\cw+\k*\dl, \y-#1+\k*\dl) rectangle (\cx+\cw+\k*\dl, \y+#1+\k*\dl);
    \draw[card] (\cx-\cw, \y-#1) rectangle (\cx+\cw, \y+#1);}
  \def\encoder#1{\node[enc] (#1) at (\ex, \y) {\textbf{Encoder}\\[-0.5pt]{\tiny\color{gray!40!black}(ModernBERT-large)}};}
  \def\output#1{\node[outp] (#1) at (\ox, \y) {$\varphi$};}

  \def\y{0}
  \begin{pgfonlayer}{background}\fill[gray!10, rounded corners=4pt] (\pl, \y+14.5) rectangle (\pr, \y-12);\end{pgfonlayer}
  \node[title] at (3, \y) {\textbf{Laya}\\{(as released)}\\[2pt]{\color{gray!40!black}$n$ passes}};
  \cards{7.6}
  \node[chq] at (\cx, \y+4.2) {question $Q_j$};
  \node[cho] at (\cx, \y)     {options $O_j$};
  \node[chi] at (\cx, \y-4.2) {instruction $\instr$};
  \node[lab] at (\cx, \y-9.7) {one forward pass per question};
  \encoder{encA}
  \node[head] (hdA) at (\hy, \y) {\textbf{Head}};
  \node[ren] (renA) at (\rx, \y) {\textbf{Renderer}};
  \output{phiA}
  \draw[arr] (\cx+\cw+2*\dl, \y) -- (encA.west);
  \draw[arr] (encA.east) -- (hdA.west);
  \draw[dashed, thin, gray!60!black, rounded corners=2pt] (24.8, \y+13) rectangle (111, \y-11.3);
  \draw[arr, mypurple!50!black] (hdA.east) -- (renA.west);
  \node[sel, below=2pt] at (108.2, \y) {$\hat{\opt}_j$};
  \node[sel, below=2pt] at (117.1, \y) {$\hat{\opt}_1,\!\dots,\!\hat{\opt}_n$};
  \draw[arr] (renA.east) -- (phiA.west);

  \def\y{-28}
  \begin{pgfonlayer}{background}\fill[gray!10, rounded corners=4pt] (\pl, \y+13) rectangle (\pr, \y-12);\end{pgfonlayer}
  \node[title] at (3, \y) {\textbf{One-pass}\\\textbf{translator}\\[2pt]{\color{gray!40!black}1 pass}};
  \draw[card] (\cx-\cw, \y-5.4) rectangle (\cx+\cw, \y+5.4);
  \node[chi] at (\cx, \y+2.1) {instruction $\instr$};
  \node[cho] at (\cx, \y-2.1) {atomic propositions};
  \encoder{encB}
  \def\hs{3.8mm}
  \coordinate (hdw) at (\hx, \y);
  \coordinate (hde) at ([xshift=\hs] hdw);
  \path[drop shadow, fill=white] ([yshift=2*\hs] hdw) rectangle ([yshift=-2*\hs] hde);
  \foreach \q [count=\i] in {$Q_1$, $Q_2$, {}, $Q_n$} {
    \node[draw=mygreen!70!black, fill=mygreen!10, thin, minimum size=\hs, text width=\hs, align=center,
      text height=1.7mm, text depth=0.4mm, inner sep=0pt, font=\tiny, anchor=north west]
      (h\i) at ([yshift=(3-\i)*\hs] hdw) {\q};
  }
  \foreach \d in {-0.8mm, 0mm, 0.8mm} \fill[black] ([yshift=\d] h3.center) circle (0.3pt);
  \node[font=\scriptsize\sffamily\bfseries, anchor=south, inner sep=1pt] at ([yshift=0.9mm] h1.north) {Heads};
  \node[ren] (renB) at (\rx, \y) {\textbf{Renderer}};
  \output{phiB}
  \draw[arr] (\cx+\cw, \y) -- (encB.west);
  \draw[arr] (encB.east) -- (hdw);
  \coordinate (join) at ([xshift=3mm] hde);
  \foreach \i in {1, 2, 4} \draw[thin, mypurple!50!black] (h\i.east) -| (join);
  \fill[mypurple!50!black] (join) circle (1.2pt);
  \draw[arr, mypurple!50!black] (join) -- node[sel, below=2pt] {$\hat{\opt}_1, \dots, \hat{\opt}_n$} (renB.west);
  \draw[arr] (renB.east) -- (phiB.west);

  \def\y{-58.5}
  \begin{pgfonlayer}{background}\fill[gray!10, rounded corners=4pt] (\pl, \y+15.5) rectangle (\pr, \y-12);\end{pgfonlayer}
  \node[title] at (3, \y) {\textbf{Level-by-level}\\\textbf{translator}\\[2pt]{\color{gray!40!black}1 pass per level}};
  \cards{7.6}
  \node[chq] at (\cx, \y+4.2) {partial formula};
  \node[cho] at (\cx, \y)     {operators, propositions};
  \node[chi] at (\cx, \y-4.2) {instruction $\instr$};
  \node[lab] at (\cx, \y-9.7) {one forward pass per level};
  \encoder{encC}
  \node[head] (hdC) at (\hy, \y) {\textbf{Head}};
  \node[ren] (renC) at (\rx, \y) {\textbf{Renderer}};
  \output{phiC}
  \draw[arr] (\cx+\cw+2*\dl, \y) -- (encC.west);
  \draw[arr] (encC.east) -- (hdC.west);
  \draw[arr, mypurple!50!black] (hdC.east) -- node[sel, below=2pt] {one symbol\\per position} (renC.west);
  \draw[arr, dashed] (renC.north) -- (\rx+7.5, \y+13.5) -- node[lab, below=1pt, pos=0.4] {partial formula (next level)}
    (\cx+2*\dl, \y+13.5) -- (\cx+2*\dl, \y+7.6+2*\dl);
  \draw[arr] (renC.east) -- (phiC.west);
\end{tikzpicture}
\caption{\textbf{Three System-One pipelines} from an NL instruction $\instr$ to a formula $\varphi$ of a fragment
$\Frac(\AP)$. In each, an encoder reads
input sequences, heads select options, and a renderer turns the selected options $\hat{\opt}$ into $\varphi$.}
\label{fig:translator}
\vspace{-4mm}%
\end{figure}

This section introduces System-One models for translation and compares them with generative language models
(Section~\ref{sec:why}). Used naively, assuming that a catalog of questions fully characterizes the fragment $\Frac(\AP)$, similar to those in Figure~\ref{fig:questions}, a System-One model needs one forward pass (i.e., model call) per question, so the catalog's size is the main cost driver. We propose two
translators whose number of forward passes is instead set in advance by design
(Figure~\ref{fig:translator}). If a sufficient\footnote{We formalize such sufficiency through the notion of
\emph{completeness} in Definition~\ref{def:complete}.} fixed question catalog exists, the \emph{one-pass translator}
answers all its questions in a single forward pass using a modified architecture (Section~\ref{sec:architecture}). If no sufficient question catalog
exists, e.g., for more complex NL requirements that can only be captured by nested LTL formulas, the
\emph{level-by-level translator} in Section~\ref{sec:levels} builds the formula hierarchically.
Section~\ref{sec:range} establishes guarantees about the accuracy of our translators.

\subsection{System-One Models for Translation}
\label{sec:why}
A \emph{System-One model} \citep{jev2026,laya2026} is a text-based
classifier: it reads a text instruction together with a question and its candidate answers (often called \emph{options}) and
returns a predicted probability for each option. It consists of an \emph{encoder}, a large neural network that
turns the input text into numerical vectors, and a \emph{head}, a small network that turns these vectors into one
probability per option (App.~\ref{app:so} gives the technical details).
It does not output a formula by itself.

To translate an instruction \seginstr{$\instr$}, a System-One model answers questions about it by selecting the most
probable option for each question. A fixed function, the \emph{renderer} $r\colon \mathbb{O} \to \LTL(\AP)$,
maps the selected options $\hat{\opt} \in \mathbb{O}$ to the output formula $T(\instr) = r(\hat{\opt})$, where
$\mathbb{O}$ is the set of all combinations of options. The three pipelines in Figure~\ref{fig:translator} follow
this scheme but differ in their questions, renderers, and required numbers of forward passes (i.e., model
calls).
We use \laya \citep{laya2026} as the base System-One model because it is open-weight and small enough to run on
local hardware, unlike the cloud-hosted model \jev (Section~\ref{sec:intro}). As released, \laya is a general-purpose classifier not trained on temporal logic, so all three pipelines require fine-tuning (Appendix~\ref{app:data}).

A generative language model (e.g., an LLM) instead produces a formula one token at a time.
A System-One model has two advantages for translation. First, it selects only allowed options, so the renderer
produces a well-formed formula by design, whereas a \emph{generated} formula is guaranteed to be well-formed only if
decoding is constrained by a grammar~\citep{english2025}. Second, the number of forward passes follows a different
cost regime. A generative language model needs one forward pass per generated output token, bounded only by an output-length
cutoff, which truncates the formula if reached. Used naively, a System-One model needs one forward pass per question in a fixed question catalog, so the question catalog's size is the main cost driver
(Section~\ref{sec:architecture}). Our \emph{one-pass translator} needs a single forward pass regardless of the question
catalog's size; our \emph{level-by-level translator} needs at most $L$ forward passes, one per level, for a maximum depth
$L$ chosen in advance. The number of forward passes is thus fixed (one-pass) or bounded (level-by-level) by design, and
independent of both the question catalog's size and the number of output tokens.

\subsection{The One-Pass Translator}
\label{sec:architecture}
The one-pass translator uses a \emph{question catalog}, a fixed set of questions
\segquestion{$Q_1,\ldots,Q_n$} about an instruction \seginstr{$\instr$}.
A question has either \emph{fixed options},
with $O_j$ the same for every instruction, or \emph{instruction-dependent options} $O_j = O_j(\seginstr{$\instr$})$. For the running example, the four questions in Figure~\ref{fig:questions} suffice. Two ask whether a pattern (visit, avoid) is present, with the fixed options \segopt{yes} and \segopt{no}, each; two ask
which places these patterns relate to, with the instruction-dependent options \segopt{street\_5} and
\segopt{parking\_lot\_2}, which the instruction text \seginstr{$\instr$} mentions.

\label{sec:question_catalogs}%
The question catalog is designed for the system's fragment $\Frac(\AP)$, either by an expert of the system or from
existing collections of specification patterns \citep{dwyer1999,menghi2021}, such as \emph{visit} and \emph{avoid}
(Appendix~\ref{app:patterns} details the specification patterns we use). For each selected pattern, the catalog asks whether the instruction \seginstr{\instr} contains it
(\segquestion{Must the robot never go to a certain place?}), how many times, and which atomic propositions it refers
to. For example, \seginstr{visit street 5, then house 4} contains two consecutive visits, whose places, i.e., atomic propositions, are
\segopt{street\_5} and \segopt{house\_4}. The atomic propositions are either fixed by the system in advance, e.g., the places of a known map,
or extracted automatically from the instruction via a template-based rule, turning \seginstr{street 5} into \segopt{street\_5}
\citep[as in, e.g.,][]{hsiung2022,chen2023,liu2023}. Appendices~\ref{app:robot} and~\ref{app:mine} give the catalogs of our experiments.

For a question catalog, the option combinations are $\mathbb{O} = O_1 \times \dots \times O_n$, and the patterns
define the renderer. Each
pattern comes with a fixed formula, such as $\Diamond\,\tau$ for \segopt{visit}, where $\tau$ is the sequence of
places to reach, and $\Box\,\neg a$ for \segopt{avoid}, where $a$ is the atomic proposition to avoid
(Table~\ref{tab:patterns}). The renderer thus depends on the catalog: these formulas are given by the same expert, or
taken from the same pattern collections, which state each pattern in LTL. 
For each pattern that the instruction \seginstr{\instr} contains, the renderer instantiates a subformula with the
selected atomic propositions. In our design, the output formula $\varphi$ is the conjunction of these subformulas because an
instruction with several requirements asks for all of them. Other connectives (e.g., disjunction) can occur \emph{within} a
pattern's formula, which can be made arbitrarily complex. For the pattern
\seginstr{visit street 5, then house 4}, selecting $\tau$ to be \segopt{street\_5}, followed by \segopt{house\_4}, yields the subformula $\Diamond(\texttt{street\_5} \wedge \Diamond\,\texttt{house\_4})$; with no additional patterns contained in \seginstr{\instr}, this is the final output formula $\varphi$. Formulas not
expressible as conjunctions of catalog patterns lie outside the translator's range of expressible formulas
(made more formal in Section~\ref{sec:range}) and are left to the level-by-level translator of Section~\ref{sec:levels}.
Answers about patterns that the instruction does not contain are ignored.

Used naively (Figure~\ref{fig:translator}, top row), a System-One model answers the question catalog one question at
a time. For each question, the encoder reads the question, its options, and the instruction, and the head selects
the most probable option. Translation thus takes $n$ forward passes.

The one-pass translator (Figure~\ref{fig:translator}, middle row) removes this dependence. It keeps \laya's encoder and replaces its head
(top row) with heads that we tailored and trained for NL to LTL translation.
Since a translator's questions \segquestion{$Q_1,\ldots,Q_n$} are fixed in advance, we move
them from the input into the model: each question \segquestion{$Q_j$} has a head of its own, which selects an option $\hat{\opt}_j$ among that question's options $O_j$,
so the single input sequence contains only the instruction \seginstr{$\instr$} and the options that depend on it (i.e., its candidate atomic propositions). The encoder reads this
sequence once and all $n$ heads read its output, so one forward pass yields the selected options $\hat{\opt} = (\hat{\opt}_1, \dots, \hat{\opt}_n)$, whatever the number $n$ of questions.
Appendix~\ref{app:speed} gives the architecture of our heads and the input sequence.

The one-pass translator is trained starting from \laya's weights, first fine-tuned on catalog questions: the encoder and the heads are fine-tuned together via cross-entropy on
instructions paired with their intended answers to the catalog's questions.\footnote{These answers can be easily obtained by reading them off programmatically from the reference formula.}Appendix~\ref{app:data} details data and training.

\subsection{The Level-by-Level Translator}
\label{sec:levels}
A question catalog covers only formulas anticipated by its patterns. In general, however, temporal logic patterns can nest to arbitrary depth.
For example, \seginstr{Whenever you are at street 5, stay out of parking lot 2 until you reach house 4} asks for
$\Box(\texttt{street\_5} \rightarrow (\neg\,\texttt{parking\_lot\_2} \,\mathcal{U}\, \texttt{house\_4}))$,
an \emph{avoid until} pattern inside a \emph{reaction} pattern; a fixed question catalog would need a separate pattern for every such
combination, and their number grows rapidly with the nesting depth. For target fragments that include such deeply nested formulas, a complete
question catalog is impractical, so neither the naive nor the one-pass translator applies. We therefore propose the
\emph{level-by-level translator} (Figure~\ref{fig:translator}, bottom row). It follows the scheme of
Section~\ref{sec:why} with the same encoder and a single head, but uses no question catalog or patterns. Instead, it
constructs the formula's syntax tree, which is not known in advance, from the root downward.

The translator repeatedly asks \emph{``which symbol goes here?''} at each \emph{open position} of a partial
formula. Its options are the fragment's temporal and Boolean operators and the instruction's atomic propositions.
Initially, the formula is a single open position. The partial formula replaces the question in the input
sequence, and one forward pass answers all open positions at the current level. The renderer only places the
selected symbols and opens positions for their arguments: an atomic proposition closes its position, a unary
operator opens one position below it, and a binary operator opens two. The next forward pass answers the new
positions; translation stops when no positions remain open. For the instruction above, five forward passes select
\segopt{$\Box$}; then \segopt{$\rightarrow$}; then \segopt{street\_5} and \segopt{$\mathcal{U}$}; then
\segopt{$\neg$} and \segopt{house\_4}; then \segopt{parking\_lot\_2}.
Every position receives exactly one symbol, and every operator receives exactly as many arguments as it accepts.
Thus, every combination of selections yields a well-formed formula, provided that translation terminates.
We ensure termination by fixing a bounded maximum depth $L>0$ and offering only atomic propositions at the final
level $L$,
which closes every remaining open position. Translation therefore takes at most $L$ forward passes.
Appendix~\ref{app:speed} details the input sequence and gives a worked example and Algorithm~\ref{alg:levels}.

The level-by-level translator requires fine-tuning because \laya as released is not trained to select symbols
in partial formulas. Unlike in Section~\ref{sec:architecture}, instruction--formula pairs need not first be
converted into answers to a catalog's questions. A training example pairs an input sequence to the model (i.e., a partial formula, the options, and the instruction) with the target symbol.
Each symbol of a reference formula yields one example: the partial formula is the reference formula as built so far,
i.e., what the translator sees at inference if all its earlier selections were right, and the target is the reference
formula's symbol at the queried position. For the formula above, the partial formula \texttt{\#} has target
\segopt{$\Box$}; $\Box\,\texttt{\#}$ has target \segopt{$\rightarrow$}; $\Box(\texttt{\#} \rightarrow \texttt{?})$
has target \segopt{street\_5}; and so on, for all seven symbols, where \texttt{\#} is the queried position and
\texttt{?} denotes other open positions.
Starting from \laya's weights, the encoder
and the head are fine-tuned together via cross-entropy on these targets (Appendix~\ref{app:data}).
At inference, earlier selections can be wrong, and such an error carries over to the levels below it.

\subsection{Range and Accuracy}
\label{sec:range}
Since the renderer maps every possible
combination of options to a valid formula, every output of our translator is syntactically correct by construction (Section~\ref{sec:problem}).

Syntactical correctness does not imply that the rendered formula $\hat{\Phi} := T(\instr)$ is accurate. In fact, accuracy depends on the trained heads and on the
distribution of $(\Instr, \Phi)$, and can only be estimated (Section~\ref{sec:experiments}). What \emph{can} be determined by the design, is the set
of formulas that $T$ can output, i.e., its \emph{range} $\Frac_T(\AP) := \{ r(\opt) \colon \opt \in \mathbb{O} \}$, where $\mathbb{O}$ is the set of option combinations and $r$ the renderer of $T$, introduced earlier.
The range $\Frac_T(\AP)$ is a
fragment of LTL, but need not match the system's fragment $\Frac(\AP)$ in general. In other words, $\Frac_T(\AP)$ may contain formulas that no user intends and it may miss formulas that users do intend. A translator that misses none
is said to be \emph{complete}:
\begin{definition}[Completeness]
  \label{def:complete}
  A translator $T$ is \emph{complete} for a fragment $\Frac(\AP)$ if for every formula $\varphi \in \Frac(\AP)$
  there is a formula $\hat{\varphi} \in \Frac_T(\AP)$ such that $\hat{\varphi} \equivL \varphi$.
\end{definition}
A translator $T$ can only be accurate if the
intended formula is among those it can output, just as a multiple-choice question can only be answered correctly if the
right answer is among its options. To make this precise, let $E_{\Phi}:=(\exists\, \hat{\Phi} \in \Frac_T(\AP) \colon \hat{\Phi} \equivL \Phi)$ be the event that the intended formula $\Phi$ has an equivalent $\hat{\Phi}\in\Frac_T(\AP)$ in the range; the following result is straightforward.
\begin{proposition}[Translation Accuracy]\label{prop:accuracy}
  If $\Prob(E_{\Phi}) > 0$, the accuracy of $T$ is $\Prob(T(\Instr) \equivL \Phi) = \Prob(E_{\Phi})\, \Prob(T(\Instr) \equivL \Phi \mid E_{\Phi})
  \leq \Prob(E_{\Phi})$.
\end{proposition}
Proposition~\ref{prop:accuracy} splits translation accuracy across completeness and a learning-based error.
Unlike $\Prob(T(\Instr) \equivL \Phi \mid E_{\Phi})$, the completeness probability $\Prob(E_{\Phi})$ depends solely on the question catalog and on the intended formulas. Either the translator is complete (i.e., $\Prob(E_{\Phi}) = 1$), i.e., the fragment $\Frac_T(\AP)$ is defined based on known
patterns; or it is incomplete, e.g., when the fragment is known only through example formulas. In the latter case, both probabilities can be estimated empirically based on $N$ instruction--formula pairs if the intended formulas of $N_E\leq N$
have an equivalent in the range $\Frac_T(\AP)$, and $M\leq N_E$ of them are translated correctly, then $\Prob(E_{\Phi}) \approx N_E / N$ and
$\Prob(T(\Instr) \equivL \Phi \mid E_{\Phi}) \approx M / N_E$. A low $\Prob(E_{\Phi})$ calls for a larger catalog; a low $\Prob(T(\Instr) \equivL \Phi \mid E_{\Phi})$
for more or better training data.

For the level-by-level translator with a maximum of $L>0$ iterations, by induction over the
syntax tree it is complete for any nested formula up to a depth of $L$. This comes at the price of up to $L$ forward passes of the encoder, instead of one. In our experiments, we cap the depth of a formula at $L=16$ (Appendix~\ref{app:speed}). Note that in practice intended formulas rarely become this deep.

Proposition~\ref{prop:accuracy} also tells which of our two translators to use. If a catalog covers the
fragment, i.e., $\Prob(E_{\Phi}) = 1$ by construction, or $\Prob(E_{\Phi})$ is close to 1 when estimated from example formulas, the
one-pass translator loses little or no accuracy to incompleteness and answers in a single forward pass. Otherwise, the
level-by-level translator, which is complete up to depth $L$, avoids this loss at the cost of up to $L$ forward passes.

\section{Experiments}
\label{sec:experiments}

To validate our translators' capabilities on local hardware, we run every translator on two machines: an older laptop with
an Apple M1 Pro chip and 16~GB of RAM, and a Raspberry Pi~5 (\Rpi) with 4~GB of RAM. For fair comparison, every translator has the same interface, with the instruction and the options as input and a
formula as output; only the prompts of the off-the-shelf and cloud-based general-purpose models also contain rules and examples
(Appendix~\ref{app:data}). All trained translators use the same training data, which
excludes all test pairs (Appendix~\ref{app:data}), and translation accuracy is measured using the same evaluation code. All code, including the Mine Rover benchmark, will be made publicly available upon acceptance.

\subsection{Benchmarks}
\label{sec:benchmarks}
We apply our System-One translator and a range of generative baselines to three benchmarks, chosen to test them across different combinations of LTL fragments, systems, and datasets.

\paragraph{Open APs.} This benchmark tests a system whose atomic propositions (APs) are not known in advance. The
dataset of \cite{wang2025} comprises 678 pairs of an instruction $\instr$ and a reference formula for the robot of the
running example, which visits locations, picks up and puts down objects, and takes photos. Instructions name numbered
locations and objects, from which the atomic propositions are extracted. The fragment is known only through the reference formulas, so we designed
the catalog after inspecting them; it uses eight patterns of Table~\ref{tab:patterns} (see Appendix~\ref{app:robot}).
These patterns render an equivalent of 674 of the 678 reference formulas, which estimates completeness $\Prob(E)$
(Proposition~\ref{prop:accuracy}) at 99.4\%. Since the catalog covers almost all reference formulas, we use
the one-pass translator (Section~\ref{sec:architecture}), which answers it in a single forward pass.

\paragraph{Unseen Patterns.} This benchmark tests whether a translator generalizes to formula shapes absent from
its training data. \cite{bayat2026} pair English sentences with LTL formulas over five colors as atomic propositions
and all eleven temporal operators of Section~\ref{sec:prelim-ltl}. We split the data by formula template, i.e., the formula with its
atomic propositions abstracted, so no test formula shares its operator structure with a training formula. The test set has
1,992 instruction--formula pairs, which comprises eight different wordings of 249 unique formulas
(App.~\ref{app:data}).
Since the test formulas have nested shapes absent from the training formulas, no fixed catalog designed from the
training data covers them, so we use the level-by-level translator (Section~\ref{sec:levels}), at the cost of one
forward pass per level of the formula; its options are the eleven operators and the colors in the sentence.

\begin{wrapfigure}{r}{74mm}
\centering
\resizebox{72mm}{!}{\input{figs/mine_rover_map}}
\vspace{0.5mm}
\refstepcounter{figure}\label{fig:mine}
\parbox{74mm}{Figure~\thefigure: The Mine Rover executing {\footnotesize\seginstr{Fetch a charge, blast vein 2,
get out of the blast zone within 60 seconds, then return to the exit}}.}
\end{wrapfigure}
\paragraph{Application to a Mine Rover.} 
The last benchmark targets NL commands for a wheeled mine rover, where local translation matters
(Figure~\ref{fig:mine}). For this, we define a fragment that lets the rover inspect an underground mine offline and on battery power. Commands conjoin up to three
instances of eleven pattern families, each a group of related templates, including system-specific ones such as a
blasting procedure (App.~\ref{app:mine}); the catalog is thus complete by construction, and we use the
one-pass translator (Section~\ref{sec:architecture}). The system
fixes all 36 atomic propositions, so the options do not depend on the command. The test set consists of 599 commands
paraphrased by a language model, the commands closest to what a human operator would type; Appendix~\ref{app:results_all}
reports results on two further test sets.

\subsection{Baseline Translators}
\label{sec:translators}
We compare our System-One translators with local generative language models, since language models are
the most accurate translators available (Section~\ref{sec:intro}).

The accuracy of the 20 \textbf{off-the-shelf models} we tested is consistently much lower (Appendix~\ref{app:data} and Table~\ref{tab:results_full}), so we fine-tune \tool{Qwen2.5} with 0.5, 1.5, and 3 billion
parameters and \tool{Llama3.2} with 1 and 3 billion parameters on the training data of each benchmark using low-rank
adaptation (LoRA) \citep{hu2022}, as is common practice \citep{pan2023,chen2023,quansah2026,ping2026,ye2026}.
The \textbf{fine-tuned models} receive a short prompt containing only the instruction and the
options. All local language models run in \tool{llama.cpp} with weights quantized to 4 bits, which is the common choice for local
deployment on a fixed memory budget~\citep{dettmers2023}.

Without fine-tuning, only large \textbf{general-purpose models} in the cloud reach usable accuracy. For
reference, we report one run each of two of them, \tool{Claude Opus 5.5} and \tool{GPT-6 Astra}, through their command-line
interfaces (Appendix~\ref{app:data}), since several translators in the literature rely on such models
(Appendix~\ref{app:related_work}). They have an unfair advantage: they run in a remote data center.

\subsection{Results}\label{sec:results}

We evaluate each methods by measuring the \emph{accuracy}, i.e. the fraction of test pairs whose output is equivalent ($\equivL$) to the reference formula
(Section~\ref{sec:prelim-ltl}), checked using \tool{Spot~2.16}~\citep{spot}; an output that does not parse counts as wrong.
\emph{Runtime} is the wall-clock time of one translation, from the instruction and its options to the rendered
formula. \emph{Energy} is the energy measured over a run, including loading the model, divided by the number of
instructions; \tool{CodeCarbon~3.3.1}~\citep{codecarbon} measures it on the M1 Pro, and the board's power-management chip on
the \Rpi{} (Appendix~\ref{app:measurement}). Every
configuration runs three times, with three training seeds for the trained translators; we report the mean and the
standard deviation over the three runs.
On the \Rpi, every translator runs on the four CPU cores, and we run all
fine-tuned language models that fit into its 4~GB at 4 bits (up to about 3B parameters).

\begin{table}[tb]
\centering
\scriptsize
\setlength{\tabcolsep}{2.2pt}
\resizebox{\textwidth}{!}{\begin{tabular}{@{}l r@{\hspace{4pt}}rr@{\hspace{8pt}}rr r@{\hspace{4pt}}rr@{\hspace{8pt}}rr r@{\hspace{4pt}}rr@{\hspace{8pt}}rr@{}}
\toprule
 & \multicolumn{5}{c}{Open APs~\citep{wang2025}} & \multicolumn{5}{c}{Unseen Patterns~\citep{bayat2026}} & \multicolumn{5}{c}{Mine Rover (New)} \\
\cmidrule(lr){2-6}\cmidrule(lr){7-11}\cmidrule(l){12-16}
 & \multicolumn{3}{c}{M1} & \multicolumn{2}{c}{\Rpi} & \multicolumn{3}{c}{M1} & \multicolumn{2}{c}{\Rpi} & \multicolumn{3}{c}{M1} & \multicolumn{2}{c}{\Rpi} \\
Translator & Acc. & Time & Energy & Time & Energy & Acc. & Time & Energy & Time & Energy & Acc. & Time & Energy & Time & Energy \\
\midrule
System-One (Ours) & 80.0$_{\pm 1.0}$ & \textbf{0.055} & \textbf{0.84} & \textbf{0.87} & \textbf{5.24} & \textbf{66.3}$_{\pm 0.9}$ & 0.27 & 4.84 & 7.47 & 54.80 & \textbf{95.4}$_{\pm 0.6}$ & \textbf{0.062} & \textbf{0.92} & \textbf{1.31} & \textbf{9.11} \\
\midrule
\multicolumn{16}{@{}l}{\textit{Fine-tuned local language models}} \\
\tool{Qwen2.5-0.5B} & 74.6$_{\pm 2.4}$ & 0.14 & 2.09 & 1.08 & 7.91 & 43.4$_{\pm 4.3}$ & \textbf{0.11} & \textbf{1.65} & \textbf{0.88} & \textbf{6.50} & 74.9$_{\pm 3.1}$ & 0.29 & 4.98 & 3.21 & 24.00 \\
\tool{Qwen2.5-1.5B} & \textbf{80.1}$_{\pm 1.0}$ & 0.26 & 5.56 & 2.24 & 15.29 & 52.2$_{\pm 2.1}$ & 0.20 & 4.15 & 1.90 & 12.92 & 83.6$_{\pm 2.4}$ & 0.59 & 12.77 & 6.37 & 44.17 \\
\tool{Qwen2.5-3B} & 77.3$_{\pm 2.7}$ & 0.43 & 10.34 & 4.60 & 31.31 & 47.4$_{\pm 1.2}$ & 0.34 & 7.66 & 3.69 & 25.51 & 79.7$_{\pm 8.4}$ & 1.01 & 24.21 & 12.68 & 89.93 \\
\tool{Llama3.2-1B} & 70.6$_{\pm 8.8}$ & 0.18 & 4.13 & 1.80 & 12.25 & 43.2$_{\pm 2.1}$ & 0.14 & 3.16 & 1.50 & 10.28 & 79.2$_{\pm 5.2}$ & 0.41 & 9.47 & 4.77 & 33.91 \\
\tool{Llama3.2-3B} & 77.3$_{\pm 0.6}$ & 0.43 & 10.07 & 4.86 & 32.70 & 40.0$_{\pm 1.3}$ & 0.35 & 8.18 & 4.05 & 27.64 & 68.4$_{\pm 7.4}$ & 1.04 & 24.95 & 13.72 & 95.48 \\
\midrule
\multicolumn{16}{@{}l}{\textit{Cloud references (not local; accuracy only)}} \\
\tool{Claude Opus 5.5} & 91.0 & -- & -- & -- & -- & 74.0 & -- & -- & -- & -- & 97.2 & -- & -- & -- & -- \\
\tool{GPT-6 Astra} (\tool{Codex}, low) & 88.5 & -- & -- & -- & -- & 73.3 & -- & -- & -- & -- & 96.2 & -- & -- & -- & -- \\
\bottomrule
\end{tabular}
}
\caption{\textbf{Results.} Accuracy in percent (mean$_{\pm\text{standard deviation}}$ over three runs and, if
applicable, three training seeds), runtime in seconds, and measured energy in Joules per instruction (including loading
the model), on the M1 Pro and the \Rpi{} (Appendix~\ref{app:results_all}). --: not applicable.
Off-the-shelf local language models are listed in Table~\ref{tab:results_full}.}
\label{tab:results}
\vspace{-4mm}%
\end{table}

\begin{figure}[t]
\centering
\includegraphics[width=\textwidth]{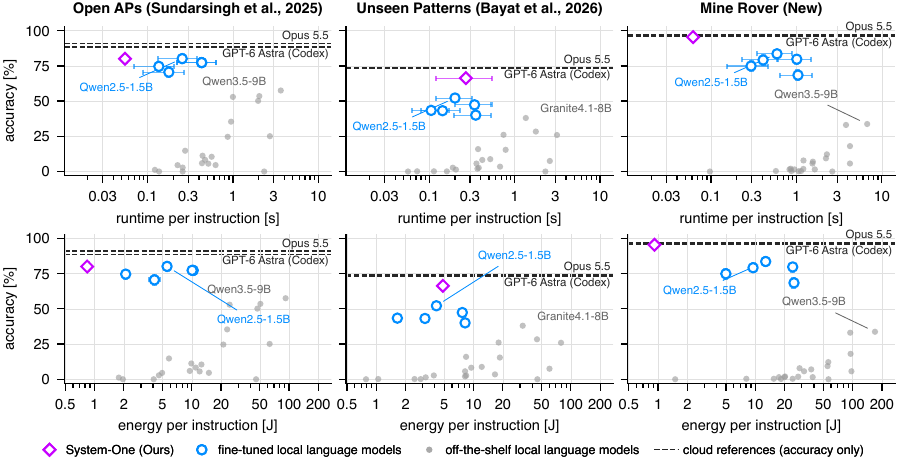}
\caption{Accuracy against runtime (top) and measured energy (bottom) per instruction on the M1 Pro. Vertical
bars: bootstrap 95\% confidence interval of the accuracy; horizontal bars: 10th to 90th percentile of the
per-instruction runtimes (top) and $\pm$ one standard deviation over the runs (bottom). Off-the-shelf models have no bars; only the most accurate model of each class is named; cloud references are
drawn at their accuracy.}
\label{fig:tradeoff}
\vspace{-4mm}%
\end{figure}

On every benchmark, our System-One translator is at least as accurate as the most accurate fine-tuned local
language model, \tool{Qwen2.5-1.5B} (Table~\ref{tab:results} and Figure~\ref{fig:tradeoff}): equally accurate on Open APs,
and 14 and 12 percent more accurate on the unseen formula shapes of Unseen Patterns and the paraphrased commands of the Mine Rover. Only
the cloud models are more accurate, and they do not run locally. Where one forward pass suffices (Open APs and the
Mine Rover), our one-pass translator also needs 5 to 10 times less runtime and 7 to 14 times less energy than
\tool{Qwen2.5-1.5B}; our level-by-level translator (Unseen Patterns), which needs several passes, costs slightly more.

On the \Rpi, the one-pass translator produces the same formulas as on the M1 Pro, needs
about a second per instruction, and keeps its advantage over \tool{Qwen2.5-1.5B}, with 2.6 to 4.9 times less runtime and
energy (Appendix~\ref{app:results_all}).
On Unseen Patterns, our level-by-level translator also produces the same formulas on the \Rpi{} and remains the most
accurate local translator (66.3 vs.\ 52.2\%), but it needs 3.9 times the runtime and 4.2 times the energy of
\tool{Qwen2.5-1.5B} (7.47 vs.\ 1.90~s and 54.8 vs.\ 12.9~J), because its several passes over long rows are limited by
arithmetic on a CPU.

\section{Conclusion and Future Work}
\label{sec:discussion_and_future_work}

Building on the new class of System-One models, we introduced the first System-One translators from NL into LTL: a one-pass and a level-by-level translator. The former translates in a single pass but only when a fixed question catalog characterizing the fragment is available. The latter applies to arbitrary nested formulas, but its cost grows with the depth of the formula. In both cases, every output is a well-formed formula by design, and the number of forward passes, and thus runtime, per instruction is bounded in advance. Our experiments suggest that they offer a better trade-off between accuracy and cost than every local LLM we tested.

System-One models have only been introduced recently; we may thus expect computational improvements in the near future. Moreover, we see four directions for further work. First, the choice of a \emph{minimal} question catalog is an interesting direction. Second, we aim to build a large-scale dataset of human-written instructions, as they are now scarce and limited to the fragment and atomic propositions of a specific system. Third, catalogs could be derived automatically from example formulas. Fourth, translation could be extended to other temporal logics, such as STL.

\subsection*{Generative AI Use}
Generative AI tools were used for coding and experimental implementation, brainstorming, as well as assisting with writing and editing the text. The authors reviewed and verified all AI-assisted content and take responsibility for the final content of the paper.

\bibliography{references}

\begin{thebibliography}{45}
\providecommand{\natexlab}[1]{#1}
\providecommand{\url}[1]{\texttt{#1}}
\expandafter\ifx\csname urlstyle\endcsname\relax
  \providecommand{\doi}[1]{doi: #1}\else
  \providecommand{\doi}{doi: \begingroup \urlstyle{rm}\Url}\fi

\bibitem[Almeida(2026)]{jev2026}
Diogo Almeida.
\newblock Introducing {System One Models} \& {Jev}.
\newblock TypeSafe AI blog, \url{https://typesafe.ai/blog/introducing-system-one-models-and-jev}, 2026.
\newblock 15 September 2026; accessed 16 October 2026.

\bibitem[Andresel et~al.(2026)Andresel, Mateis, Nickovic, Kounoupidis, Katsaros, and Tripakis]{andresel2026}
Medina Andresel, Cristinel Mateis, Dejan Nickovic, Spyridon Kounoupidis, Panagiotis Katsaros, and Stavros Tripakis.
\newblock Ltlguard: Formalizing ltl specifications with compact language models and lightweight symbolic reasoning.
\newblock arXiv:2603.05728, 2026.

\bibitem[Baier and Katoen(2008)]{baier2008}
Christel Baier and Joost-Pieter Katoen.
\newblock \emph{Principles of Model Checking}.
\newblock MIT Press, Cambridge, MA, 2008.
\newblock ISBN 978-0-262-02649-9.

\bibitem[Barrientos et~al.(2025)Barrientos, Winter, and Rinderle-Ma]{barrientos2025}
Marisol Barrientos, Karolin Winter, and Stefanie Rinderle-Ma.
\newblock Automatic extraction and formalization of temporal requirements from text: A survey.
\newblock In \emph{Enterprise Design, Operations, and Computing}, pages 259--278, 2025.
\newblock \doi{10.1007/978-3-031-78338-8_14}.

\bibitem[Bayat et~al.(2026)Bayat, Ozay, Abate, and Jungers]{bayat2026}
Amir Bayat, Necmiye Ozay, Alessandro Abate, and Raphael~M. Jungers.
\newblock {End-to-End} {Abstraction-Based} control with {LLM-Enhanced} {NL-to-LTL} translation.
\newblock \emph{IEEE Control Systems Letters}, pages 1--1, 2026.
\newblock \doi{10.1109/lcsys.2026.3738589}.

\bibitem[Belta et~al.(2017)Belta, Yordanov, and Aydin~Gol]{belta2017}
Calin Belta, Boyan Yordanov, and Ebru Aydin~Gol.
\newblock \emph{Formal Methods for Discrete-Time Dynamical Systems}, volume~89 of \emph{Studies in Systems, Decision and Control}.
\newblock Springer, Cham, 2017.

\bibitem[Bombieri et~al.(2023)Bombieri, Meli, Dall'Alba, Rospocher, and Fiorini]{bombieri2023}
Marco Bombieri, Daniele Meli, Diego Dall'Alba, Marco Rospocher, and Paolo Fiorini.
\newblock Mapping natural language procedures descriptions to linear temporal logic templates: an application in the surgical robotic domain.
\newblock \emph{Applied Intelligence}, 53\penalty0 (22):\penalty0 26351--26363, 2023.
\newblock \doi{10.1007/s10489-023-04882-0}.

\bibitem[Brunello et~al.(2019)Brunello, Montanari, and Reynolds]{brunello2019}
Andrea Brunello, Angelo Montanari, and Mark Reynolds.
\newblock Synthesis of {LTL} formulas from natural language texts: State of the art and research directions.
\newblock In \emph{26th International Symposium on Temporal Representation and Reasoning (TIME 2019)}, volume 147 of \emph{Leibniz International Proceedings in Informatics (LIPIcs)}, pages 17:1--17:19, 2019.
\newblock \doi{10.4230/LIPIcs.TIME.2019.17}.

\bibitem[Buzhinsky(2019)]{buzhinsky2019}
Igor Buzhinsky.
\newblock Formalization of natural language requirements into temporal logics: a survey.
\newblock In \emph{2019 IEEE 17th International Conference on Industrial Informatics (INDIN)}, pages 400--406, 2019.
\newblock \doi{10.1109/indin41052.2019.8972130}.

\bibitem[Chen et~al.(2023)Chen, Gandhi, Zhang, and Fan]{chen2023}
Yongchao Chen, Rujul Gandhi, Yang Zhang, and Chuchu Fan.
\newblock {NL2TL}: Transforming natural languages to temporal logics using large language models.
\newblock In \emph{Proceedings of the 2023 Conference on Empirical Methods in Natural Language Processing}, pages 15880--15903, 2023.
\newblock \doi{10.18653/v1/2023.emnlp-main.985}.

\bibitem[Chen et~al.(2024)Chen, Arkin, Dawson, Zhang, Roy, and Fan]{chen2024}
Yongchao Chen, Jacob Arkin, Charles Dawson, Yang Zhang, Nicholas Roy, and Chuchu Fan.
\newblock {AutoTAMP}: Autoregressive task and motion planning with {LLMs} as translators and checkers.
\newblock In \emph{2024 IEEE International Conference on Robotics and Automation (ICRA)}, pages 6695--6702, 2024.
\newblock \doi{10.1109/icra57147.2024.10611163}.

\bibitem[{Convai Innovations}(2026)]{laya2026}
{Convai Innovations}.
\newblock {Laya}: A non-autoregressive {System 1} decision engine.
\newblock \url{https://github.com/NandhaKishorM/laya}, 2026.
\newblock Open-source software (Apache-2.0), version 0.3.27; models at \url{https://huggingface.co/convaiinnovations/laya}; accessed 16 October 2026.

\bibitem[Cosler et~al.(2023)Cosler, Hahn, Mendoza, Schmitt, and Trippel]{cosler2023}
Matthias Cosler, Christopher Hahn, Daniel Mendoza, Frederik Schmitt, and Caroline Trippel.
\newblock nl2spec: Interactively translating unstructured natural language to temporal logics with large language models.
\newblock In \emph{Computer Aided Verification}, pages 383--396, 2023.
\newblock \doi{10.1007/978-3-031-37703-7_18}.

\bibitem[Courty et~al.(2026)Courty, Schmidt, Luccioni, Feld, et~al.]{codecarbon}
Benoit Courty, Victor Schmidt, Sasha Luccioni, Boris Feld, et~al.
\newblock {CodeCarbon}: Estimate and track carbon emissions from computing.
\newblock Software, version 3.3.1, \url{https://codecarbon.io}, 2026.

\bibitem[Dettmers and Zettlemoyer(2023)]{dettmers2023}
Tim Dettmers and Luke Zettlemoyer.
\newblock The case for 4-bit precision: k-bit inference scaling laws.
\newblock In \emph{Proceedings of the 40th International Conference on Machine Learning (ICML)}, volume 202 of \emph{Proceedings of Machine Learning Research}, pages 7750--7774, 2023.

\bibitem[Devlin et~al.(2019)Devlin, Chang, Lee, and Toutanova]{devlin2019}
Jacob Devlin, Ming-Wei Chang, Kenton Lee, and Kristina Toutanova.
\newblock {BERT}: Pre-training of deep bidirectional transformers for language understanding.
\newblock In \emph{Proceedings of the 2019 Conference of the North {A}merican Chapter of the {A}ssociation for {C}omputational {L}inguistics: Human Language Technologies (Volume 1: Long and Short Papers)}, pages 4171--4186, Minneapolis, Minnesota, 2019. Association for Computational Linguistics.
\newblock \doi{10.18653/v1/N19-1423}.

\bibitem[Duret-Lutz et~al.(2022)Duret-Lutz, Renault, Colange, Renkin, Gbaguidi~Aisse, Schlehuber-Caissier, Medioni, Martin, Dubois, Gillard, and Lauko]{spot}
Alexandre Duret-Lutz, Etienne Renault, Maximilien Colange, Florian Renkin, Alexandre Gbaguidi~Aisse, Philipp Schlehuber-Caissier, Thomas Medioni, Antoine Martin, J{\'e}r{\^o}me Dubois, Cl{\'e}ment Gillard, and Henrich Lauko.
\newblock From {S}pot 2.0 to {S}pot 2.10: What's new?
\newblock In \emph{Proceedings of the 34th International Conference on Computer Aided Verification (CAV)}, volume 13372 of \emph{Lecture Notes in Computer Science}, pages 174--187. Springer, 2022.
\newblock \doi{10.1007/978-3-031-13188-2_9}.

\bibitem[Dwyer et~al.(1999)Dwyer, Avrunin, and Corbett]{dwyer1999}
Matthew~B. Dwyer, George~S. Avrunin, and James~C. Corbett.
\newblock Patterns in property specifications for finite-state verification.
\newblock In \emph{Proceedings of the 21st international conference on Software engineering}, ICSE99, pages 411--420, 1999.
\newblock \doi{10.1145/302405.302672}.

\bibitem[English et~al.(2025)English, Simon, Jha, and Ewetz]{english2025}
William English, Dominic Simon, Sumit~Kumar Jha, and Rickard Ewetz.
\newblock Grammar-forced translation of natural language to temporal logic using {LLMs}.
\newblock In \emph{Proceedings of the 42nd International Conference on Machine Learning (ICML)}, volume 267 of \emph{Proceedings of Machine Learning Research}, 2025.
\newblock arXiv:2512.16814.

\bibitem[Finucane et~al.(2010)Finucane, Jing, and Kress-Gazit]{finucane2010}
C~Finucane, Gangyuan Jing, and H~Kress-Gazit.
\newblock {LTLMoP}: Experimenting with language, temporal logic and robot control.
\newblock In \emph{2010 IEEE/RSJ International Conference on Intelligent Robots and Systems}, pages 1988--1993, 2010.
\newblock \doi{10.1109/iros.2010.5650371}.

\bibitem[Firoozi et~al.(2025)Firoozi, Tucker, Tian, Majumdar, Sun, Liu, Zhu, Song, Kapoor, Hausman, Ichter, Driess, Wu, Lu, and Schwager]{firoozi2025}
Roya Firoozi, Johnathan Tucker, Stephen Tian, Anirudha Majumdar, Jiankai Sun, Weiyu Liu, Yuke Zhu, Shuran Song, Ashish Kapoor, Karol Hausman, Brian Ichter, Danny Driess, Jiajun Wu, Cewu Lu, and Mac Schwager.
\newblock Foundation models in robotics: Applications, challenges, and the future.
\newblock \emph{The International Journal of Robotics Research}, 44\penalty0 (5):\penalty0 701--739, 2025.
\newblock \doi{10.1177/02783649241281508}.

\bibitem[Fuggitti and Chakraborti(2023)]{fuggitti2023}
Francesco Fuggitti and Tathagata Chakraborti.
\newblock {NL2LTL} -- a python package for converting natural language ({NL}) instructions to linear temporal logic ({LTL}) formulas.
\newblock \emph{Proceedings of the AAAI Conference on Artificial Intelligence}, 37\penalty0 (13):\penalty0 16428--16430, 2023.
\newblock \doi{10.1609/aaai.v37i13.27068}.

\bibitem[Giannakopoulou et~al.(2020)Giannakopoulou, Pressburger, Mavridou, and Schumann]{giannakopoulou2020}
Dimitra Giannakopoulou, Thomas Pressburger, Anastasia Mavridou, and Johann Schumann.
\newblock Generation of formal requirements from structured natural language.
\newblock In \emph{Requirements Engineering: Foundation for Software Quality}, pages 19--35, 2020.
\newblock \doi{10.1007/978-3-030-44429-7_2}.

\bibitem[Gopalan et~al.(2018)Gopalan, Arumugam, Wong, and Tellex]{gopalan2018}
Nakul Gopalan, Dilip Arumugam, Lawson Wong, and Stefanie Tellex.
\newblock {Sequence-to-Sequence} language grounding of {Non-Markovian} task specifications.
\newblock In \emph{Robotics: Science and Systems XIV}, RSS2018, 2018.
\newblock \doi{10.15607/rss.2018.xiv.067}.

\bibitem[Hahn et~al.(2022)Hahn, Schmitt, Tillman, Metzger, Siber, and Finkbeiner]{hahn2022}
Christopher Hahn, Frederik Schmitt, Julia~J. Tillman, Niklas Metzger, Julian Siber, and Bernd Finkbeiner.
\newblock Formal specifications from natural language.
\newblock arXiv:2206.01962, 2022.

\bibitem[He et~al.(2022)He, Bartocci, Ničković, Isakovic, and Grosu]{he2022}
Jie He, Ezio Bartocci, Dejan Ničković, Haris Isakovic, and Radu Grosu.
\newblock {DeepSTL}: from english requirements to signal temporal logic.
\newblock In \emph{Proceedings of the 44th International Conference on Software Engineering}, ICSE '22, pages 610--622, 2022.
\newblock \doi{10.1145/3510003.3510171}.

\bibitem[Hsiung et~al.(2022)Hsiung, Mehta, Chu, Liu, Patel, Tellex, and Konidaris]{hsiung2022}
Eric Hsiung, Hiloni Mehta, Junchi Chu, Xinyu Liu, Roma Patel, Stefanie Tellex, and George Konidaris.
\newblock Generalizing to new domains by mapping natural language to lifted {LTL}.
\newblock In \emph{2022 International Conference on Robotics and Automation (ICRA)}, pages 3624--3630, 2022.
\newblock \doi{10.1109/icra46639.2022.9812169}.

\bibitem[Hu et~al.(2022)Hu, Shen, Wallis, Allen-Zhu, Li, Wang, Wang, and Chen]{hu2022}
Edward~J. Hu, Yelong Shen, Phillip Wallis, Zeyuan Allen-Zhu, Yuanzhi Li, Shean Wang, Lu~Wang, and Weizhu Chen.
\newblock {LoRA}: Low-rank adaptation of large language models.
\newblock In \emph{International Conference on Learning Representations (ICLR)}, 2022.
\newblock arXiv:2106.09685.

\bibitem[Kehoe et~al.(2015)Kehoe, Patil, Abbeel, and Goldberg]{kehoe2015}
Ben Kehoe, Sachin Patil, Pieter Abbeel, and Ken Goldberg.
\newblock A survey of research on cloud robotics and automation.
\newblock \emph{IEEE Transactions on Automation Science and Engineering}, 12\penalty0 (2):\penalty0 398--409, 2015.
\newblock \doi{10.1109/TASE.2014.2376492}.

\bibitem[Kress-Gazit et~al.(2008)Kress-Gazit, Fainekos, and Pappas]{kressgazit2008}
Hadas Kress-Gazit, Georgios~E. Fainekos, and George~J. Pappas.
\newblock Translating structured english to robot controllers.
\newblock \emph{Advanced Robotics}, 22\penalty0 (12):\penalty0 1343--1359, 2008.
\newblock \doi{10.1163/156855308x344864}.

\bibitem[Li and Zhao(2025)]{t3planner2025}
Jia Li and Guoxiang Zhao.
\newblock T3 planner: A self-correcting llm framework for robotic motion planning with temporal logic.
\newblock arXiv:2510.16767, 2025.

\bibitem[Lignos et~al.(2014)Lignos, Raman, Finucane, Marcus, and Kress-Gazit]{lignos2015}
Constantine Lignos, Vasumathi Raman, Cameron Finucane, Mitchell Marcus, and Hadas Kress-Gazit.
\newblock Provably correct reactive control from natural language.
\newblock \emph{Autonomous Robots}, 38\penalty0 (1):\penalty0 89--105, 2014.
\newblock \doi{10.1007/s10514-014-9418-8}.

\bibitem[Liu et~al.(2023)Liu, Yang, Idrees, Liang, Schornstein, Tellex, and Shah]{liu2023}
Jason~Xinyu Liu, Ziyi Yang, Ifrah Idrees, Sam Liang, Benjamin Schornstein, Stefanie Tellex, and Ankit Shah.
\newblock Grounding complex natural language commands for temporal tasks in unseen environments.
\newblock In \emph{Proceedings of the 7th Conference on Robot Learning (CoRL)}, volume 229 of \emph{Proceedings of Machine Learning Research}, 2023.
\newblock Lang2LTL; arXiv:2302.11649.

\bibitem[Ma et~al.(2025)Ma, Wen, Su, Liang, Tian, Qin, and Yang]{ma2025}
Zhi Ma, Cheng Wen, Zhexin Su, Xiao Liang, Cong Tian, Shengchao Qin, and Mengfei Yang.
\newblock Bridging natural language and formal {Specification--Automated} translation of software requirements to {LTL} via hierarchical semantics decomposition using {LLMs}.
\newblock In \emph{2025 40th IEEE/ACM International Conference on Automated Software Engineering (ASE)}, pages 1208--1220, 2025.
\newblock \doi{10.1109/ase63991.2025.00104}.

\bibitem[Menghi et~al.(2021)Menghi, Tsigkanos, Pelliccione, Ghezzi, and Berger]{menghi2021}
Claudio Menghi, Christos Tsigkanos, Patrizio Pelliccione, Carlo Ghezzi, and Thorsten Berger.
\newblock Specification patterns for robotic missions.
\newblock \emph{IEEE Transactions on Software Engineering}, 47\penalty0 (10):\penalty0 2208--2224, 2021.
\newblock \doi{10.1109/tse.2019.2945329}.

\bibitem[Newcomb and Ochoa(2026)]{newcomb2026}
Alexandra Newcomb and Omar Ochoa.
\newblock Automatic translation of unstructured requirements into linear temporal logic through large language models.
\newblock arXiv:2608.06287, 2026.

\bibitem[Nikora and Balcom(2009)]{nikora2009}
Allen~P. Nikora and Galen Balcom.
\newblock Automated identification of {LTL} patterns in natural language requirements.
\newblock In \emph{2009 20th International Symposium on Software Reliability Engineering}, pages 185--194, 2009.
\newblock \doi{10.1109/issre.2009.15}.

\bibitem[Pan et~al.(2023)Pan, Chou, and Berenson]{pan2023}
Jiayi Pan, Glen Chou, and Dmitry Berenson.
\newblock {Data-Efficient} learning of natural language to linear temporal logic translators for robot task specification.
\newblock In \emph{2023 IEEE International Conference on Robotics and Automation (ICRA)}, pages 11554--11561, 2023.
\newblock \doi{10.1109/icra48891.2023.10161125}.

\bibitem[Ping et~al.(2026)Ping, Ding, Liang, Zhou, Lei, Chen, Wu, Zhou, and Zhang]{ping2026}
Yuqi Ping, Huahao Ding, Tianhao Liang, Longyu Zhou, Guangyu Lei, Xinglin Chen, Junwei Wu, Jieyu Zhou, and Tingting Zhang.
\newblock {LLM-Enabled} {Low-Altitude} {UAV} natural language navigation via signal temporal logic specification translation and repair.
\newblock \emph{IEEE Transactions on Cognitive Communications and Networking}, 12:\penalty0 11218--11232, 2026.
\newblock \doi{10.1109/tccn.2026.3717937}.

\bibitem[Quansah and Bonnah(2026)]{quansah2026}
Paapa~Kwesi Quansah and Ernest Bonnah.
\newblock {NeuroNL2LTL}: A neurosymbolic framework for natural language translation of linear temporal logic.
\newblock In \emph{Formal Techniques for Distributed Objects, Components, and Systems}, pages 56--73, 2026.
\newblock \doi{10.1007/978-3-032-28187-6_4}.

\bibitem[Rozier(2016)]{rozier2016}
Kristin~Yvonne Rozier.
\newblock Specification: The biggest bottleneck in formal methods and autonomy.
\newblock In Sandrine Blazy and Marsha Chechik, editors, \emph{Verified Software. Theories, Tools, and Experiments (VSTTE 2016)}, volume 9971 of \emph{Lecture Notes in Computer Science}, pages 8--26. Springer, 2016.
\newblock \doi{10.1007/978-3-319-48869-1_2}.

\bibitem[Sundarsingh et~al.(2025)Sundarsingh, Wang, Deshmukh, and Kantaros]{wang2025}
David~Smith Sundarsingh, Jun Wang, Jyotirmoy~V. Deshmukh, and Yiannis Kantaros.
\newblock Conformalnl2ltl: Translating natural language instructions into temporal logic formulas with conformal correctness guarantees.
\newblock arXiv:2504.21022, 2025.

\bibitem[Warner et~al.(2025)Warner, Chaffin, Clavi{\'e}, Weller, Hallstr{\"o}m, Taghadouini, Gallagher, Biswas, Ladhak, Aarsen, Adams, Howard, and Poli]{warner2025}
Benjamin Warner, Antoine Chaffin, Benjamin Clavi{\'e}, Orion Weller, Oskar Hallstr{\"o}m, Said Taghadouini, Alexis Gallagher, Raja Biswas, Faisal Ladhak, Tom Aarsen, Griffin~Thomas Adams, Jeremy Howard, and Iacopo Poli.
\newblock Smarter, better, faster, longer: A modern bidirectional encoder for fast, memory efficient, and long context finetuning and inference.
\newblock In \emph{Proceedings of the 63rd Annual Meeting of the {A}ssociation for {C}omputational {L}inguistics (Volume 1: Long Papers)}, pages 2526--2547, Vienna, Austria, 2025. Association for Computational Linguistics.
\newblock \doi{10.18653/v1/2025.acl-long.127}.

\bibitem[Ye et~al.(2026)Ye, Li, Huang, Ma, and Yin]{ye2026}
Bowen Ye, Zhijian Li, Junyue Huang, Junkai Ma, and Xiang Yin.
\newblock Reasonstl: Bridging natural language and signal temporal logic via tool-augmented process-rewarded learning.
\newblock arXiv:2605.06483, 2026.

\bibitem[Zrelli et~al.(2026)Zrelli, Misson, Attia, de~Magalhaes, Shabah, and Nicolescu]{zrelli2026}
Rim Zrelli, Henrique~Amaral Misson, Marwa~Ben Attia, Felipe~Gohring de~Magalhaes, Abdo Shabah, and Gabriela Nicolescu.
\newblock Automatic translation of natural language requirements into {CTL} specifications using large language models: A multi-approach evaluation.
\newblock \emph{Journal of Systems and Software}, 240:\penalty0 112941, 2026.
\newblock \doi{10.1016/j.jss.2026.112941}.

\end{thebibliography}

\appendix

\section{Related Work}\label{app:related_work}
Translators from natural language into temporal logic have gone through three generations, reviewed below;
surveys cover the field in more detail \citep{brunello2019,buzhinsky2019,barrientos2025}. To our knowledge, none of
the works below measures energy or compares accuracy and runtime on the same hardware, and none uses a System-One
model.

\paragraph{Rule-, Grammar-, and Pattern-Based Translators.}
These translators parse structured or controlled English \citep{kressgazit2008,finucane2010,lignos2015,giannakopoulou2020,menghi2021}.
Some of them already treat translation as classification: they classify an instruction into specification
patterns and fill in their arguments, with feature-based classifiers \citep{nikora2009}, semantic role labeling
\citep{bombieri2023}, intent and entity classifiers \citep{fuggitti2023}, or a BERT pattern classifier followed by a
large language model \citep{zrelli2026}. Their classifiers are trained for one task each.

\paragraph{Trained Neural Translators.}
These translators learn the mapping from pairs of sentences and formulas \citep{gopalan2018,hahn2022,he2022,hsiung2022,pan2023}.

\paragraph{Translators Based on Large Language Models.}
Large language models, with few-shot prompting, decomposition, verification loops, or fine-tuning, are now the
most common approach \citep{cosler2023,chen2023,liu2023,english2025,wang2025,bayat2026} and outperform earlier
neural translators on shared benchmarks \citep{pan2023,chen2023,liu2023}.

\paragraph{Local Translators.}
Closest to our work are translators that run entirely on local,
open-weight models, fine-tuned or prompted, from a few hundred million to 14 billion parameters
\citep{pan2023,english2025,andresel2026,quansah2026,ping2026,ye2026}; a few state local deployment as a goal
\citep{andresel2026,t3planner2025}.

\section{From \laya to Our Translators}
\label{app:so}\label{app:speed}
This appendix gives the technical details that Section~\ref{sec:method} omits: the input sequence and encoder of
\laya as released, and the three changes that turn it into our translators, namely the input sequence and heads of the
one-pass translator, the recognition of pattern families on the Mine Rover, and the input sequences of the
level-by-level translator. None of these changes modifies the encoder.

\paragraph{\laya as Released.}
\laya answers choice questions, yes-or-no questions, and scores on a scale; our translators use only the first
two, and a yes-or-no question has the two options \segopt{yes} and \segopt{no}. For each question, \laya reads one
input sequence that combines the \segtype{type} of the question, the \segquestion{question}, its
\segopt{options {$\opt_1, \dots, \opt_m$}}, and the \seginstr{instruction}:
\begin{equation}
  \text{\small
  \segshadowtrue
  \mbox{%
  \segsp{[CLS]} \segtype{$\langle$type$\rangle$:} \segquestion{$\langle$Q$\rangle$} \segsp{[SEP]} \segsp{[MASK]}
  \segopt{$\langle\opt_1\rangle$} $\cdots$ \segsp{[MASK]} \segopt{$\langle\opt_m\rangle$} \segsp{[SEP]} \seginstr{$\langle$\instr$\rangle$}
  \segsp{[SEP]}}}\label{eq:encoding_laya}
\end{equation}
The gray tokens are the special tokens of BERT-style encoders \citep{devlin2019}: \segsp{[CLS]} opens the
sequence, \segsp{[SEP]} closes each segment, and a \segsp{[MASK]} token precedes each option and marks it (its
\emph{marker}). The encoder, \tool{ModernBERT-large} \citep{warner2025}, returns a vector, an \emph{output state}, for
every token. Its attention is bidirectional, so each output state depends on the whole input sequence; the output
state at the marker of an option thus represents that option in the context of the question, the instruction, and the
other options. \laya's head scores the output state at each marker, and a softmax over the markers gives the
probability of each option.

\paragraph{One-Pass Translator.}
The instruction-dependent options of a catalog are atomic propositions that the instruction can refer to, its
\emph{candidates} $\mathcal{C}(\instr) = \{c_1,\ldots,c_k\}$. They are either extracted from the instruction by a
hand-written rule (Appendix~\ref{app:robot}) or fixed by the system (the 20 places of the Mine Rover; Appendix~\ref{app:mine}).
The single input sequence contains the instruction and each candidate once:
  \begin{equation*}
    \text{\small
    \segshadowtrue
    \segsp{[CLS]} \seginstr{$\langle$\instr$\rangle$} \segsp{[SEP]}
    $\text{\segsp{[MASK]} \segopt{$\langle c_1\rangle$} $\cdots$ \segsp{[MASK]}
    \segopt{$\langle c_k\rangle$}}$
    \segsp{[SEP]}}
  \end{equation*}
Unlike in \eqref{eq:encoding_laya}, the instruction comes first; since attention is bidirectional and the
translator is fine-tuned on sequences of this form, the order matters little. Our heads read two kinds of output
states: $h_{\mathrm{CLS}} \in \R^{1024}$, the output state of \segsp{[CLS]}, which depends on the whole input sequence
and serves as its summary, and $h_c \in \R^{1024}$, the output state at the marker of candidate $c$, which represents
$c$ in the context of the instruction. The heads are trained from scratch and are small: for the catalog of Open APs, about 1.7 million parameters together, against 421 million in the encoder.
\begin{itemize}
  \item \emph{Fixed options.} All questions with fixed options share one feed-forward network on
  $h_{\mathrm{CLS}}$, $z = W_2\, \mathrm{GELU}\big(W_1\, \mathrm{LN}(h_{\mathrm{CLS}}) + b_1\big) + b_2$,
  where $\mathrm{LN}$ is a layer normalization and $W_1 \in \R^{1024 \times 1024}$. The output $z$ has one entry per
  option of every such question (388 entries for the 76 fixed-option questions of Open APs); question $Q_j$ takes its
  own slice $z_{O_j}$, and a softmax over this slice gives the probability $p_j(\opt \mid \instr)$ of each option
  $\opt \in O_j$. The meaning of each entry is learned in training, so fixed options need no tokens in the input
  sequence.
  \item \emph{Instruction-dependent options.} For each candidate $c \in \mathcal{C}(\instr)$, $h_c$ is projected
  to $m_c = P\, \mathrm{LN}(h_c) \in \R^{256}$. Question $Q_j$ is represented
  by a learned embedding $q_j \in \R^{256}$, the sum of embeddings of its pattern, instance, task, task kind, and slot
  (for example, the destination of the second task of an \emph{eventually} clause). The score of candidate $c$ for $Q_j$
  is $s_j(c) = w^\top \mathrm{GELU}(m_c + q_j) + b$, and a softmax over the candidates allowed for $Q_j$,
  $O_j(\instr) \subseteq \mathcal{C}(\instr)$, gives $p_j(c \mid \instr)$; a question about a place, for example,
  considers only places. $P$, $w$, and $b$ are shared by all such questions, which differ only in $q_j$.
\end{itemize}

\paragraph{Pattern Families on the Mine Rover.}
On the Mine Rover, the presence of each pattern family is not read from $h_{\mathrm{CLS}}$: a score is computed
for every token of the command and maximized over these tokens. A family is thus recognized from the words that
indicate it most strongly, regardless of which other families occur with it, including in combinations held out of
training.

\paragraph{Level-by-Level Translator.}
Each input sequence contains the partial formula in place of a question, with ``\texttt{\#}'' at the queried position
and ``\texttt{?}'' at the other open positions, followed by the options (the fragment's operators, written as words, and
the atomic propositions named in the instruction) and the instruction. For example, the formula of
\seginstr{Visit street 5, then house 4, and never enter parking lot 2} is
$\Diamond(\texttt{street\_5} \wedge \Diamond\,\texttt{house\_4}) \wedge \Box\,\neg\,\texttt{parking\_lot\_2}$. The
first three passes select \segopt{$\wedge$}; then \segopt{$\Diamond$} and \segopt{$\Box$}; then \segopt{$\wedge$} and \segopt{$\neg$}. The fourth pass reads a batch
of three input sequences:
\begin{equation*}
  \text{\scriptsize
  \segshadowtrue
  $\left.\begin{tabular}{@{}l@{}}
  \segsp{[CLS]} \segquestion{$\Diamond(\texttt{\#}\wedge\texttt{?})\wedge\Box\,\neg\,\texttt{?}$} \segsp{[SEP]}\\[3pt]
  \segsp{[CLS]} \segquestion{$\Diamond(\texttt{?}\wedge\texttt{\#})\wedge\Box\,\neg\,\texttt{?}$} \segsp{[SEP]}\\[3pt]
  \segsp{[CLS]} \segquestion{$\Diamond(\texttt{?}\wedge\texttt{?})\wedge\Box\,\neg\,\texttt{\#}$} \segsp{[SEP]}
  \end{tabular}\right\}$
  $\begin{tabular}{@{}l@{}}
  \segsp{[MASK]} \segopt{not} \segsp{[MASK]} \segopt{next} $\cdots$ \segsp{[MASK]} \segopt{weak until} \;\textcolor{gray}{$\hookleftarrow$}\\[3pt]
  \segsp{[MASK]} \segopt{street 5} \segsp{[MASK]} \segopt{house 4} \segsp{[MASK]} \segopt{parking lot 2} \segsp{[SEP]} \;\textcolor{gray}{$\hookleftarrow$}\\[3pt]
  \seginstr{Visit street 5, then house 4, and never enter parking lot 2} \segsp{[SEP]}
  \end{tabular}$}
\end{equation*}
and selects \segopt{street\_5}, \segopt{eventually}, and \segopt{parking\_lot\_2}; the fifth pass selects
\segopt{house\_4}. Only atomic propositions are offered at depth 16, or once the formula could exceed 64 symbols, which
closes the formula instead of truncating it. Algorithm~\ref{alg:levels} summarizes the translator.

\paragraph{Empty Selections.} If the selected options determine no part of a formula at all (with a catalog: no
pattern is present; level by level: no atomic proposition is offered), the renderer returns $\top$; this choice is
arbitrary and serves only to ensure that every output is syntactically correct. In none of our benchmarks is $\top$
an intended formula.

\begin{algorithm2e}[h]
\caption{Level-by-level Translation}
\label{alg:levels}
\KwIn{instruction $\instr$; options $O$ comprising the operators of the fragment and the atomic propositions named in
$\instr$}
\KwOut{formula $\hat{\varphi}$, where $|\hat{\varphi}|$ denotes the number of symbols in $\hat{\varphi}$}
\lIf{$O \cap \AP = \emptyset$}{\Return $\top$}
$\hat{\varphi} \gets$ a single open root position; $P \gets \{\text{root}\}$\;
\For{$d = 1, \dots, 16$, while $P \neq \emptyset$}{
  $O_d \gets O$\;
  \lIf(\tcp*[f]{atomic propositions only}){$d = 16$ or $|\hat{\varphi}| + 3|P| > 64$}{$O_d \gets O \cap \AP$}
  \ForEach{$p \in P$}{
    $s_p \gets$ the input sequence consisting of $\hat{\varphi}$ with $p$ marked by \texttt{\#} and all other positions in
    $P$ by \texttt{?}, followed by $O_d$ with operators written as words, and then the instruction $\instr$\;
  }
  Run one encoder forward pass on the batch $\{s_p\}_{p \in P}$\;
  \ForEach{$p \in P$}{
    $\hat{\opt}_p \gets$ the most probable option in $O_d$ for $s_p$\;
    Place $\hat{\opt}_p$ at $p$ in $\hat{\varphi}$\;
  }
  $P \gets$ the new open child positions of the chosen symbols, with zero children for an atomic proposition, one for
  a unary operator, and two for a binary operator\;
}
\Return $\hat{\varphi}$\;
\end{algorithm2e}

\section{Question Catalogs}
\label{app:catalogs}
This appendix gives the shared pattern library from which our catalogs are drawn, then the catalogs of
the Open APs robot and of the Mine Rover (Section~\ref{sec:benchmarks}).

\subsection{Pattern Library}
\label{app:patterns}
A \emph{specification pattern} is a requirement that recurs across systems, with an LTL template whose slots
are filled with atomic propositions. Our templates follow \citet{dwyer1999}, in their global scope, and the 22 patterns
that \citet{menghi2021} derived for robotic missions. Table~\ref{tab:patterns} lists our shared library, with the
corresponding pattern of each source, where one exists.

\begin{table}[h]
\centering
\footnotesize
\begin{tabular}{@{}l>{$}l<{$}ll@{}}
\toprule
\textbf{Pattern} & \text{\textbf{Template}} & \textbf{\citet{dwyer1999}} & \textbf{\citet{menghi2021}} \\
\midrule
visit          & \Diamond\,\tau                              & existence              & (sequenced) visit \\
repeated visit & \Box\Diamond\,\tau                          & --                     & sequenced patrolling \\
patrol         & \Box\Diamond\,\ell                          & --                     & patrolling \\
avoid          & \Box\,\neg a                                & absence                & global avoidance \\
stay           & \Box\,\ell                                  & universality           & -- \\
reach and stay & \Diamond\Box\,\ell                          & --                     & -- \\
wait           & \ell \,\mathcal{U}\, \tau                   & --                     & wait \\
avoid until    & \neg a \,\mathcal{U}\, \tau                 & precedence$^{\dagger}$ & past avoidance \\
reaction       & \Box(e \rightarrow \Diamond\,\tau),\ \Box(e \rightarrow a) & response$^{\ddagger}$ & delayed, instantaneous reaction \\
\bottomrule
\end{tabular}
\caption{Shared pattern library. Each catalog uses a subset of these patterns, possibly with system-specific
ones (Appendices~\ref{app:robot} and~\ref{app:mine}); formulas that no pattern covers are left to the level-by-level
translator (Appendix~\ref{app:speed}). $\ell$ is a location, $a$ an atomic proposition, and $e$ a sensor reading set
by the environment. $\tau$ is a sequence of steps to be reached one after another (Section~\ref{sec:architecture}). $^{\dagger}$With weak until in \citet{dwyer1999}. $^{\ddagger}$Only
$\Box(e \rightarrow \Diamond\,\tau)$.}
\label{tab:patterns}
\end{table}

\subsection{Open APs Robot}
\label{app:robot}
\paragraph{Atomic Propositions.} The robot's atomic propositions represent locations, such as \texttt{street\_5}, which hold when the robot is there; objects, such as
\texttt{p\_box\_4}, which hold when it picks them up; and two actions, \texttt{photo} (taking a photo) and \texttt{pd}
(putting down the carried object). Instructions can name 46 kinds of location and 22 kinds of object, each with a
number. A hand-written rule converts each known numbered name in the
instruction $\instr$, such as \seginstr{street 5}, into an atomic proposition and adds the two actions. For the running
example, it yields \texttt{street\_5}, \texttt{parking\_lot\_2}, \texttt{photo}, and \texttt{pd}. These options
contain every atomic proposition in the reference formula for 98.2\% of the pairs; misses result from typos and
inconsistent names in the data.

\paragraph{Catalog.} The catalog uses the eight patterns of Table~\ref{tab:patterns} other than reaction, with up to three
instances each. A step sequence $\tau$ comprises up to three tasks of eight kinds: go to a location; take a
photo, pick up an object, or put down the carried object at a location; deliver an object between locations; and
pick up, put down, or take a photo without a location. Each kind defines a short step sequence over its slots.
Delivering object $o$ from $\ell_1$ to $\ell_2$, for example, has steps $\ell_1$, $o$, and $\ell_2 \wedge
\texttt{pd}$. A \emph{stay} flag requires the robot to remain at the last location forever.

\paragraph{Range.} The four reference
formulas outside the catalog's range combine atomic propositions in ways that the patterns exclude, such as $\Box\,\neg(\texttt{house\_4} \wedge
\texttt{p\_bottle\_2})$.

\paragraph{Held-Out Check.} Because we designed the catalog after inspecting the reference formulas, we also evaluated
on the 452 pairs never used for design decisions: the one-pass translator reaches 79.1 $\pm$ 1.2\% there vs.
80.0 $\pm$ 1.0\% on all pairs, a difference similar to that of the most accurate fine-tuned language model,
\tool{Qwen2.5-1.5B} (78.8 $\pm$ 1.6\% vs. 80.1 $\pm$ 1.0\%).

\subsection{Mine Rover}
\label{app:mine}
The mine rover's 36 atomic propositions comprise 20 places (drifts, crosscuts, stopes, gold veins, the refuge,
the magazine, and the exit), eight actions (taking a sample, scanning, taking a photo, dropping a radio beacon,
reporting, unloading, holding still, and placing an explosive charge), three sensor readings set by the environment
(gas, gold, and a person nearby), one battery proposition (\texttt{low\_battery}, true while the rover's battery is
below 30\% of its capacity), and four blasting propositions (one clock proposition for each of three
deadlines, such as \texttt{deadline\_60}, and \texttt{blast\_zone}). The one-pass translator receives the 20 places as
options, and the actions and sensor readings as fixed options; the language models receive all 36 atomic propositions
as options. Table~\ref{tab:mine} lists the pattern families.

\begin{table}[h]
\centering
\small
\begin{tabular}{@{}ll@{}}
\toprule
\textbf{Family} & \textbf{Template} \\
\midrule
visit$^{\ast}$   & $\Diamond\,p$ or $\Diamond(p \wedge a)$ \\
ordered          & $\Diamond(p_1 \wedge a_1 \wedge \Diamond(p_2 \wedge a_2 \wedge \cdots))$, 2 to 4 places, each
                   action optional \\
patrol           & $\bigwedge_{p \in P} \Box\Diamond\,p$, $P$ 1 to 3 places or a group of places \\
avoid$^{\ast}$   & $\Box\,\neg p$ \\
stay out until   & $\neg p \,\mathcal{U}\, q$ \\
delivery         & $\Diamond(p \wedge a \wedge \Diamond(\texttt{exit} \wedge b))$ \\
response$^{\ast}$& one of eight rules, such as $\Box(\texttt{gas} \rightarrow \Diamond(\texttt{exit} \wedge
                   \texttt{report}))$ or $\Box(\texttt{person} \rightarrow \texttt{hold})$ \\
gas avoidance    & $\Box(\texttt{gas} \rightarrow (\neg p \,\mathcal{U}\, \texttt{exit}))$ \\
blasting         & $\Diamond(\texttt{explosive} \wedge s) \wedge \Box(\texttt{explosive} \rightarrow
                   (\neg\texttt{deadline\_}T \,\mathcal{U}\, \neg\texttt{blast\_zone}))$ \\
survey           & $\bigwedge_{s \in S} \Diamond(s \wedge \texttt{scan}) \wedge \Box(\texttt{scan} \rightarrow
                   \bigvee_{s \in S} s) \wedge \Box(\texttt{gold} \leftrightarrow a)$ \\
park             & $\Diamond\Box\,p$ \\
\bottomrule
\end{tabular}
\caption{The mine rover's eleven pattern families. $p$, $q$ are places, $s$ a stope, $S$ a set of stopes,
and $a$, $b$ actions. A delivery is one of (\texttt{sample}, \texttt{unload}), (\texttt{photo}, \texttt{report}), and
(\texttt{scan}, \texttt{report}); a survey takes a sample or photo where gold is detected. In blasting, $T \in
\{30, 60, 120\}$, and the robot may also have to return to the exit after placing the charge ($\Diamond(\texttt{explosive}
\wedge s \wedge \Diamond\,\texttt{exit})$). A command conjoins one to three instances; families marked `$^{\ast}$'
occur at most twice per command, and the others at most once.}
\label{tab:mine}
\end{table}

\section{Training Data and Settings}
\label{app:data}
\paragraph{Data.}
Every translator holds out its own small validation split from its training pool, and no test pair is used for
training. On Open APs, all translators are trained on the same 15,000 synthetic instructions, which we generate from
hand-written phrasings of the catalog's eight patterns; all 678 pairs of \cite{wang2025} are used for testing only.
The fine-tuned language models output formulas and are not limited to the
catalog, so we also trained them on 25,000 synthetic instructions that additionally contain two patterns outside the
catalog (a reaction to an event and a choice between tasks), to check that restricting the training data to the
catalog does not put them at a disadvantage. Their accuracy changes by between $-0.1$ and $+4.7$ percentage points (\tool{Qwen2.5-0.5B} $74.6 \pm
2.4 \to 77.0 \pm 2.8\%$, \tool{Qwen2.5-1.5B} $80.1 \pm 1.0 \to 80.0 \pm 3.1\%$, \tool{Qwen2.5-3B} $77.3 \pm 2.7 \to 80.1 \pm
1.5\%$, \tool{Llama3.2-1B} $70.6 \pm 8.8 \to 75.3 \pm 3.8\%$, \tool{Llama3.2-3B} $77.3 \pm 0.6 \to 77.2 \pm 5.3\%$; three
training seeds each), comparable to the spread over training seeds, and the highest accuracy among the five, 80.1\%,
is the same with either training set.
On Unseen Patterns, the training pool contains the 4,680 training sentences of \cite{bayat2026}, 16,000
rule-based sentences from a grammar fitted to the training formulas, and 7,488 paraphrases of the rule-based sentences
written by \tool{Claude Opus 5.5}. We keep a paraphrase only if a separate call of the same model, which sees only the paraphrase and the colors named in the sentence as options, translates it back into a formula equivalent to the original one; 7,488 of 7,780 paraphrases pass this check.
Importantly, the 4,680 training sentences do not cover the operator structures of the test
formulas. Trained on them alone, the level-by-level translator reaches $44.1 \pm 1.5\%$ over three training seeds,
and for the first seed, 96\% of its errors have a wrong operator structure.
The generated formulas include operator structures that are absent from the training sentences, but none has the
structure of a test formula or is equivalent to one, so the test still measures generalization to unseen structures;
with them, the level-by-level translator reaches $66.3 \pm 0.9\%$ (Table~\ref{tab:results}). On the Mine Rover, the
training pool contains 22,201 synthetic commands and paraphrases.

\paragraph{Training of Our Translators.}
Our translators minimize the cross-entropy
$-\sum_i \log p_i(\opt_i^\star \mid \instr)$, where $\opt_i^\star$ is the intended answer to question $Q_i$ and $p_i$
the probability that its head assigns to an option; questions that do not apply to the instruction, such as the slots
of an absent pattern, are left out of the sum.
The intended answers come with each generated instruction on Open APs and the Mine Rover, and are read off the
reference formula on Unseen Patterns (Section~\ref{sec:levels}). The encoder and the heads are trained with AdamW,
learning rates $2.5 \cdot 10^{-5}$ and $10^{-4}$, respectively, batch size 16, weight decay 0.01, and a one-cycle
schedule. One-pass translators are trained for two epochs, starting from \laya fine-tuned to answer the catalog's
questions on 2,500 of the 15,000 synthetic instructions of Open APs (no test data), as the language models start from
their instruction tuning; the level-by-level translator is trained for one
epoch, starting from \laya.
Our translators are validated by accuracy.

\paragraph{Fine-Tuned Language Models.}
Language models are fine-tuned with LoRA for one epoch, using batch size 4 (2 with gradient checkpointing
for 3B models), learning rate $10^{-4}$, and loss on the answer only. LoRA freezes the model's weights and trains small
low-rank matrices (rank 8) added to every linear map of the last 16 transformer blocks (all blocks of \tool{Llama3.2-1B});
16 is the default of the \tool{MLX} implementation, and adapting only the last blocks reduces the memory and time needed for
training. They are validated by loss. Because a language model's
training can diverge late in a run, we evaluate its validation loss every 200 steps and keep the LoRA weights saved at
each evaluation; if the final validation loss exceeds twice its lowest value, we use the weights from the step with the
lowest validation loss, as occurred in 16 of 45 runs.

\paragraph{Prompted Models.}
We prompt 20 instruction-tuned local language models with 270 million
to 9 billion parameters using rules, examples, and the options, with thinking disabled and greedy decoding. None
exceeds 57.5, 38.0, and 33.7\% on Open APs, Unseen Patterns, and the Mine Rover, respectively
(Table~\ref{tab:results_full}).
We call \tool{Claude Opus 5.5} through \tool{Claude Code} with its default reasoning effort,
and \tool{GPT-6 Astra} through \tool{Codex} with low reasoning effort; \tool{Codex} adds about 13,600 tokens of its own instructions to the
prompt.

\section{Measurement of Runtime and Energy}\label{app:measurement}
The same code measures every translator on both machines. For each run, it loads the translator, translates
every test instruction in a fixed order, and records the platform, the library versions, the CPU load, and the
thermal state. A run starts only if no other training or benchmark process is running and the machine is on mains
power, and the machine is kept awake throughout.

\paragraph{Runtime.} The runtime (Section~\ref{sec:results}) includes every model call; the options are
extracted beforehand, outside the timed call. The first three instructions of a run are warm-up and are not
counted; we report the mean over the remaining instructions. The time to load the model enters the energy, not
the runtime.

\paragraph{Energy on the M1 Pro.} \tool{CodeCarbon}~3.3.1 \citep{codecarbon} reads the power of the CPU and the
GPU from the operating system's \tool{powermetrics} tool every 2~s, a reading taking about 1.3~s, and assigns the memory a
fixed 3~W. We changed three things: we read the CPU power instead of estimating it from the processor's thermal design
power; the CPU and the GPU share one reading, instead of two overlapping \tool{powermetrics} processes that at times leave
one of them without a reading; and we compute the energy as the sum of the mean CPU, GPU, and memory powers times the
duration of the run, because \tool{CodeCarbon}'s time intervals between readings miscount the 1.3~s of each reading (for
the constant 3~W of the memory, they yield 1.9~W). Under a steady GPU load, the resulting powers agree with an
independent \tool{powermetrics} run (GPU 21.76 vs.\ 21.81~W, CPU 0.62 vs.\ 0.56~W). The energy excludes the
display, the disk, and the losses in the power adapter.

\paragraph{Energy on the Raspberry Pi.} The power-management chip of the Raspberry Pi~5 reports the current and the
voltage of each supply rail (\texttt{vcgencmd pmic\_read\_adc}). Once per second, we sum voltage times current over
all rails with both readings and integrate this power over the run with the trapezoidal rule. A reading takes about
24~ms, about 2\% of one of the four cores. The rails cover the processor and the board's input/output, and wireless, but not the memory, whose rails report no current, nor the losses in converting the 5~V input; no
USB device is attached. The board draws 2.1~W at idle and 5.0~W with all four cores busy. As on the M1 Pro, the
energy is that of the computation, not that drawn from the wall socket, so we compare energies only between
translators measured on the same machine.

\section{Results of All Translators}\label{app:results_all}
\begin{figure}[h]
\centering
\includegraphics[width=\textwidth]{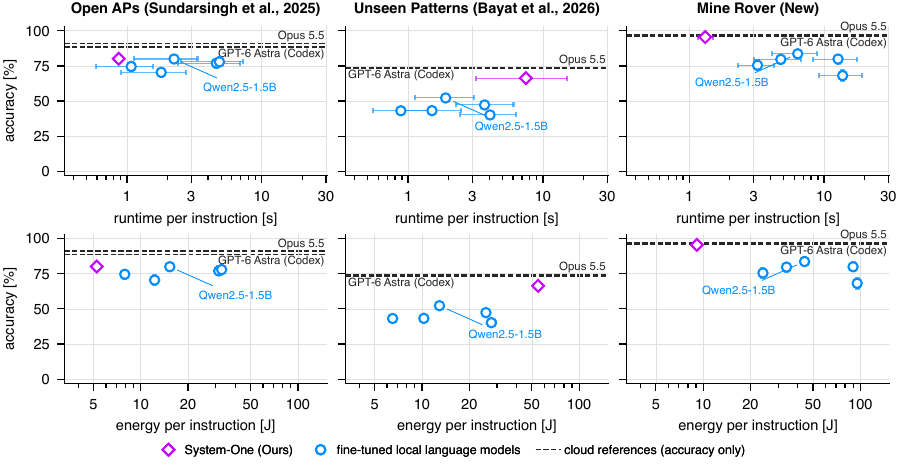}
\caption{Accuracy against runtime (top) and measured energy (bottom) per instruction on the \Rpi; three runs
(training seeds 0 to 2), with the encoding of Figure~\ref{fig:tradeoff}. Vertical bars: bootstrap 95\% confidence
interval of the accuracy over test pairs (mean of the bounds over the three runs); horizontal bars: 10th to 90th
percentile of the per-instruction runtimes (top) and $\pm$ one standard deviation over the runs (bottom). The
energy is that of the supply rails of the \Rpi{} (Appendix~\ref{app:measurement}).}
\label{fig:tradeoff_pi}
\end{figure}

Table~\ref{tab:results_full} reports every translator we tested, including those left out of
Table~\ref{tab:results}, and Figure~\ref{fig:tradeoff_pi} shows the accuracy against runtime and energy on the
\Rpi.
\begin{table}[h]
\centering
\scriptsize
\setlength{\tabcolsep}{2.2pt}
\resizebox{\textwidth}{!}{\begin{tabular}{@{}l r r@{\hspace{4pt}}rr@{\hspace{8pt}}r@{\hspace{4pt}}rr r@{\hspace{4pt}}rr@{\hspace{8pt}}r@{\hspace{4pt}}rr r@{\hspace{4pt}}rr@{\hspace{8pt}}r@{\hspace{4pt}}rr@{}}
\toprule
 & & \multicolumn{6}{c}{Open APs~\citep{wang2025}} & \multicolumn{6}{c}{Unseen Patterns~\citep{bayat2026}} & \multicolumn{6}{c}{Mine Rover (New)} \\
\cmidrule(lr){3-8}\cmidrule(lr){9-14}\cmidrule(l){15-20}
 & & \multicolumn{3}{c}{M1} & \multicolumn{3}{c}{\Rpi} & \multicolumn{3}{c}{M1} & \multicolumn{3}{c}{\Rpi} & \multicolumn{3}{c}{M1} & \multicolumn{3}{c}{\Rpi} \\
Translator & Par. & Acc. & Time & Energy & Acc. & Time & Energy & Acc. & Time & Energy & Acc. & Time & Energy & Acc. & Time & Energy & Acc. & Time & Energy \\
\midrule
System-One (Ours) & 0.40 & 80.0$_{\pm 1.0}$ & \textbf{0.055} & \textbf{0.84} & \textbf{80.0} & \textbf{0.87} & \textbf{5.24} & \textbf{66.3}$_{\pm 0.9}$ & 0.27 & 4.84 & \textbf{66.3} & 7.47 & 54.80 & \textbf{95.4}$_{\pm 0.6}$ & \textbf{0.062} & \textbf{0.92} & \textbf{95.4} & \textbf{1.31} & \textbf{9.11} \\
\midrule
\multicolumn{20}{@{}l}{\textit{Fine-tuned local language models (LoRA, direct)}} \\
\tool{Qwen2.5-0.5B} & 0.49 & 74.6$_{\pm 2.4}$ & 0.14 & 2.09 & 74.5 & 1.08 & 7.91 & 43.4$_{\pm 4.3}$ & 0.11 & 1.65 & 43.1 & \textbf{0.88} & \textbf{6.50} & 74.9$_{\pm 3.1}$ & 0.29 & 4.98 & 75.5 & 3.21 & 24.00 \\
\tool{Qwen2.5-1.5B} & 1.54 & \textbf{80.1}$_{\pm 1.0}$ & 0.26 & 5.56 & 79.9 & 2.24 & 15.29 & 52.2$_{\pm 2.1}$ & 0.20 & 4.15 & 52.2 & 1.90 & 12.92 & 83.6$_{\pm 2.4}$ & 0.59 & 12.77 & 83.6 & 6.37 & 44.17 \\
\tool{Qwen2.5-3B} & 3.09 & 77.3$_{\pm 2.7}$ & 0.43 & 10.34 & 76.9 & 4.60 & 31.31 & 47.4$_{\pm 1.2}$ & 0.34 & 7.66 & 47.3 & 3.69 & 25.51 & 79.7$_{\pm 8.4}$ & 1.01 & 24.21 & 79.9 & 12.68 & 89.93 \\
\tool{Llama3.2-1B} & 1.24 & 70.6$_{\pm 8.8}$ & 0.18 & 4.13 & 70.4 & 1.80 & 12.25 & 43.2$_{\pm 2.1}$ & 0.14 & 3.16 & 43.2 & 1.50 & 10.28 & 79.2$_{\pm 5.2}$ & 0.41 & 9.47 & 79.6 & 4.77 & 33.91 \\
\tool{Llama3.2-3B} & 3.21 & 77.3$_{\pm 0.6}$ & 0.43 & 10.07 & 77.8 & 4.86 & 32.70 & 40.0$_{\pm 1.3}$ & 0.35 & 8.18 & 40.2 & 4.05 & 27.64 & 68.4$_{\pm 7.4}$ & 1.04 & 24.95 & 68.1 & 13.72 & 95.48 \\
\midrule
\multicolumn{20}{@{}l}{\textit{Off-the-shelf local language models (direct prompting)}} \\
\tool{Gemma2-2B} & 2.61 & 11.2$_{\pm 0.0}$ & 0.44 & 9.88 & -- & -- & -- & 2.6$_{\pm 0.0}$ & 0.38 & 8.40 & -- & -- & -- & 2.0$_{\pm 0.0}$ & 0.86 & 20.17 & -- & -- & -- \\
\tool{Gemma3-1B} & 1.00 & 2.8$_{\pm 0.0}$ & 0.26 & 4.59 & -- & -- & -- & 0.1$_{\pm 0.0}$ & 0.16 & 2.86 & -- & -- & -- & 0.3$_{\pm 0.0}$ & 0.87 & 15.49 & -- & -- & -- \\
\tool{Gemma3-270M} & 0.27 & 0.0$_{\pm 0.0}$ & 0.14 & 1.96 & -- & -- & -- & 0.0$_{\pm 0.0}$ & \textbf{0.056} & \textbf{0.78} & -- & -- & -- & 0.0$_{\pm 0.0}$ & 0.097 & 1.50 & -- & -- & -- \\
\tool{Granite4.0-1B} & 1.63 & 0.0$_{\pm 0.0}$ & 2.32 & 45.75 & -- & -- & -- & 0.0$_{\pm 0.0}$ & 2.29 & 45.26 & -- & -- & -- & 0.0$_{\pm 0.0}$ & 2.63 & 56.44 & -- & -- & -- \\
\tool{Granite4.0-350M} & 0.35 & 1.2$_{\pm 0.0}$ & 0.12 & 1.78 & -- & -- & -- & 0.2$_{\pm 0.0}$ & 0.072 & 1.05 & -- & -- & -- & 0.3$_{\pm 0.0}$ & 0.58 & 8.08 & -- & -- & -- \\
\tool{Granite4.1-8B} & 8.79 & 50.1$_{\pm 0.0}$ & 1.97 & 47.18 & \multicolumn{3}{c}{\toolarge} & 38.0$_{\pm 0.0}$ & 1.35 & 32.00 & \multicolumn{3}{c}{\toolarge} & 18.0$_{\pm 0.0}$ & 4.24 & 95.28 & \multicolumn{3}{c}{\toolarge} \\
\tool{Llama3.1-8B} & 8.03 & 52.9$_{\pm 0.0}$ & 1.00 & 24.48 & \multicolumn{3}{c}{\toolarge} & 15.4$_{\pm 0.0}$ & 0.78 & 18.63 & \multicolumn{3}{c}{\toolarge} & 12.2$_{\pm 0.0}$ & 2.27 & 56.21 & \multicolumn{3}{c}{\toolarge} \\
\tool{Llama3.2-1B} & 1.24 & 4.5$_{\pm 0.1}$ & 0.23 & 4.95 & -- & -- & -- & 0.2$_{\pm 0.0}$ & 0.11 & 2.47 & -- & -- & -- & 0.3$_{\pm 0.0}$ & 1.13 & 24.18 & -- & -- & -- \\
\tool{Llama3.2-3B} & 3.21 & 5.6$_{\pm 0.1}$ & 0.52 & 12.03 & -- & -- & -- & 5.7$_{\pm 0.1}$ & 0.36 & 8.18 & -- & -- & -- & 1.8$_{\pm 0.0}$ & 1.21 & 27.55 & -- & -- & -- \\
\tool{Phi3.5-mini} & 3.82 & 25.1$_{\pm 0.0}$ & 2.71 & 63.07 & -- & -- & -- & 7.5$_{\pm 0.0}$ & 2.58 & 60.25 & -- & -- & -- & 5.7$_{\pm 0.0}$ & 4.23 & 97.17 & -- & -- & -- \\
\tool{Phi4-mini} & 3.84 & 4.6$_{\pm 0.0}$ & 0.63 & 14.98 & -- & -- & -- & 8.8$_{\pm 0.0}$ & 0.52 & 12.13 & -- & -- & -- & 1.5$_{\pm 0.0}$ & 1.64 & 38.86 & -- & -- & -- \\
\tool{Qwen2.5-0.5B} & 0.49 & 0.0$_{\pm 0.0}$ & 0.26 & 4.06 & -- & -- & -- & 1.4$_{\pm 0.0}$ & 0.21 & 3.22 & -- & -- & -- & 0.2$_{\pm 0.0}$ & 1.06 & 18.33 & -- & -- & -- \\
\tool{Qwen2.5-1.5B} & 1.54 & 14.7$_{\pm 0.0}$ & 0.28 & 5.82 & -- & -- & -- & 2.7$_{\pm 0.0}$ & 0.19 & 3.90 & -- & -- & -- & 1.5$_{\pm 0.0}$ & 0.83 & 17.48 & -- & -- & -- \\
\tool{Qwen2.5-3B} & 3.09 & 10.5$_{\pm 0.0}$ & 0.56 & 12.54 & -- & -- & -- & 15.9$_{\pm 0.0}$ & 0.37 & 8.37 & -- & -- & -- & 7.8$_{\pm 0.0}$ & 1.21 & 28.48 & -- & -- & -- \\
\tool{Qwen2.5-7B} & 7.62 & 35.4$_{\pm 0.0}$ & 0.95 & 23.07 & \multicolumn{3}{c}{\toolarge} & 26.1$_{\pm 0.0}$ & 0.74 & 18.08 & \multicolumn{3}{c}{\toolarge} & 9.3$_{\pm 0.0}$ & 2.22 & 55.22 & \multicolumn{3}{c}{\toolarge} \\
\tool{Qwen3.5-0.8B} & 0.75 & 5.9$_{\pm 0.0}$ & 0.43 & 9.53 & -- & -- & -- & 2.0$_{\pm 0.0}$ & 0.37 & 8.17 & -- & -- & -- & 1.5$_{\pm 0.0}$ & 0.95 & 20.02 & -- & -- & -- \\
\tool{Qwen3.5-2B} & 1.88 & 24.6$_{\pm 0.0}$ & 0.87 & 21.09 & -- & -- & -- & 3.5$_{\pm 0.0}$ & 0.67 & 16.60 & -- & -- & -- & 7.0$_{\pm 0.0}$ & 1.56 & 37.53 & -- & -- & -- \\
\tool{Qwen3.5-4B} & 4.21 & 53.5$_{\pm 0.0}$ & 2.04 & 50.67 & -- & -- & -- & 28.4$_{\pm 0.0}$ & 1.77 & 44.85 & -- & -- & -- & 33.1$_{\pm 0.0}$ & 3.80 & 94.45 & -- & -- & -- \\
\tool{Qwen3.5-9B} & 8.95 & 57.5$_{\pm 0.0}$ & 3.64 & 92.07 & \multicolumn{3}{c}{\toolarge} & 25.9$_{\pm 0.0}$ & 3.13 & 79.21 & \multicolumn{3}{c}{\toolarge} & 33.7$_{\pm 0.0}$ & 6.73 & 169.72 & \multicolumn{3}{c}{\toolarge} \\
\tool{SmolLM3-3B} & 3.08 & 8.3$_{\pm 0.0}$ & 0.47 & 10.97 & -- & -- & -- & 8.2$_{\pm 0.0}$ & 0.42 & 9.46 & -- & -- & -- & 5.7$_{\pm 0.0}$ & 1.54 & 31.67 & -- & -- & -- \\
\midrule
\multicolumn{20}{@{}l}{\textit{Cloud references (not local; accuracy only)}} \\
\tool{Claude Opus 5.5} & -- & 91.0 & -- & -- & -- & -- & -- & 74.0 & -- & -- & -- & -- & -- & 97.2 & -- & -- & -- & -- & -- \\
\tool{GPT-6 Astra} (\tool{Codex}, low) & -- & 88.5 & -- & -- & -- & -- & -- & 73.3 & -- & -- & -- & -- & -- & 96.2 & -- & -- & -- & -- & -- \\
\bottomrule
\end{tabular}
}
\caption{Results of every translator in the paper, with the notation of Table~\ref{tab:results} and, in addition, the
accuracy on the \Rpi; Par.: billions of parameters. Off-the-shelf models were not run on the \Rpi{}: they are much less accurate, and some do not fit into its 4~GB of
memory.}
\label{tab:results_full}
\end{table}

\paragraph{Mine Rover.} Besides the 599 paraphrased commands, the Mine Rover has two additional test sets: commands for new
formulas and commands that combine pattern families held out of training. On these, the one-pass translator reaches
98.9 and 99.3\%, and \tool{Qwen2.5-1.5B} 94.6 and 93.6\%; both are more accurate than on the paraphrases (95.4 and 83.6\%),
which are thus the harder test.

\paragraph{Raspberry Pi: Formulas and Accuracy.} The \Rpi{} columns of Table~\ref{tab:results} are the means
over the same three runs as on the M1 Pro. Our translators produce the same formulas on both machines for every
test pair in every run: the one-pass translator on Open APs and the Mine Rover, and the level-by-level translator on
Unseen Patterns. In contrast, the formulas of the fine-tuned language models agree with those of the same run on the
M1 Pro for only 84.2 to 98.8\% of the test pairs. They decode greedily, but \tool{llama.cpp} sums the same products
in a different order on the GPU of the M1 Pro and on the CPU of the \Rpi{}, which can change a token where the two most
probable ones are nearly tied, and with it the rest of the formula. For every translator, the mean accuracy over the
three runs is within 0.5 percentage points of the M1 Pro's.

\paragraph{Raspberry Pi: Runtime and Energy.} The advantage of the one-pass translator over \tool{Qwen2.5-1.5B}
is smaller on the \Rpi{} than on the M1 Pro (2.6 vs.\ 4.7 times less runtime on Open APs), likely because on a CPU the
encoder's single pass over all options is limited by arithmetic, whereas generation with a small language model is
limited by memory bandwidth. Moreover, \tool{PyTorch} has no fast 16-bit kernels for the CPU of the \Rpi{} (16-bit
weights were about 60 times slower than 32-bit ones), so the encoder runs in 32-bit there. Every translator needs 2.7
to 9.9 times more energy per instruction on the \Rpi{} than on the M1 Pro, and still 1.9 to 6.9 times more without the
idle draw of the board, even though the \Rpi{} draws less power (about 5 to 8~W against 8 to 24~W) but takes 8 to 21
times as long per instruction.

\end{document}